\documentclass[5p,times,twocolumn,authoryear]{elsarticle}

\usepackage{amssymb}
\usepackage{amsmath}
\usepackage{mathtools}
\newcommand{\R}{\mathbb{R}}

\usepackage{booktabs} 
\usepackage{physics}
\usepackage{mathtools}
\usepackage{dsfont}
\usepackage{multirow}
\usepackage{algorithm}
\usepackage{algpseudocode}
\usepackage{siunitx} 

\usepackage{dsfont}
\usepackage[hidelinks]{hyperref} 
\usepackage{orcidlink}
\usepackage{caption}
\usepackage{xcolor}
\usepackage[most]{tcolorbox}
\newcommand{\cmark}{ \textcolor{green!60!black}{\checkmark} }
\newcommand{\xmark}{ \textcolor{red!60!black}{$\times$} }
\usepackage{subcaption}
\DeclareMathOperator*{\argmax}{arg\,max}

\journal{Control Engineering Practice}

\begin{document}

\begin{frontmatter}



\title{Automated Discovery of Laser Dicing Processes with \\ Bayesian Optimization for Semiconductor Manufacturing }



\cortext[cor1]{Corresponding author.}
\cortext[cor2]{Authors contributed equally}
\author[label1]{David Leeftink\orcidlink{0000-0002-9542-3334}\corref{cor1}} 
\ead{david.leeftink@ru.nl}
\author[label2]{Roman Doll} 
\author[label2]{Heleen Visserman}
\author[label2]{Marco Post} 
\author[label2]{Faysal Boughorbel} 
\author[label1]{Max Hinne\orcidlink{0000-0002-9279-6725}\corref{cor2}} 
\author[label1]{Marcel van Gerven\orcidlink{0000-0002-2206-9098}\corref{cor2}}

\affiliation[label1]{organization={Department of Machine Learning and Neural Computing, Donders Institute for Brain, Cognition and Behaviour, Radboud University},
            city={Nijmegen},
            postcode={6525XZ},
            country={the Netherlands}}

\affiliation[label2]{organization={ASMPT Center of Competency},
            city={Beuningen},
            country={the Netherlands}}

\begin{abstract}
Laser dicing of semiconductor wafers is a critical step in microelectronic manufacturing, where multiple sequential laser passes precisely separate individual dies from the wafer. Adapting this complex sequential process to new wafer materials typically requires weeks of expert effort to balance process speed, separation quality, and material integrity.  
We present the first automated discovery of production-ready laser dicing processes on an industrial LASER1205 dicing system. We formulate the problem as a high-dimensional, constrained multi-objective Bayesian optimization task, and introduce a sequential two-level fidelity strategy to minimize expensive destructive die-strength evaluations.
On bare silicon and product wafers, our method autonomously delivers feasible configurations that match or exceed expert baselines in production speed, die strength, and structural integrity, using only technician-level operation. Post-hoc validation of different weight configurations of the utility functions reveals that multiple feasible solutions with qualitatively different trade-offs can be obtained from the final surrogate model. Expert-refinement of the discovered process can further improve production speed while preserving die strength and structural integrity, surpassing purely manual or automated methods.
\end{abstract}


\end{frontmatter}

\section{Introduction}

\begin{figure*}[t]
    \begin{tcolorbox}[
        enhanced, 
        width=1.0\linewidth, 
        boxrule=1pt,         
        colback=blue!1!white,      
        colframe=blue!50!black, 
        arc=4mm,             
        title=\color{black}{\textbf{Automated Discovery of Laser Dicing Processes}}\vspace*{1mm}, 
        fonttitle=\bfseries\Large, 
        attach title to upper=\quad,  
        top=1mm,            
        bottom=1mm,          
    ]
        \centering
       
        \includegraphics[width=1.0\linewidth]{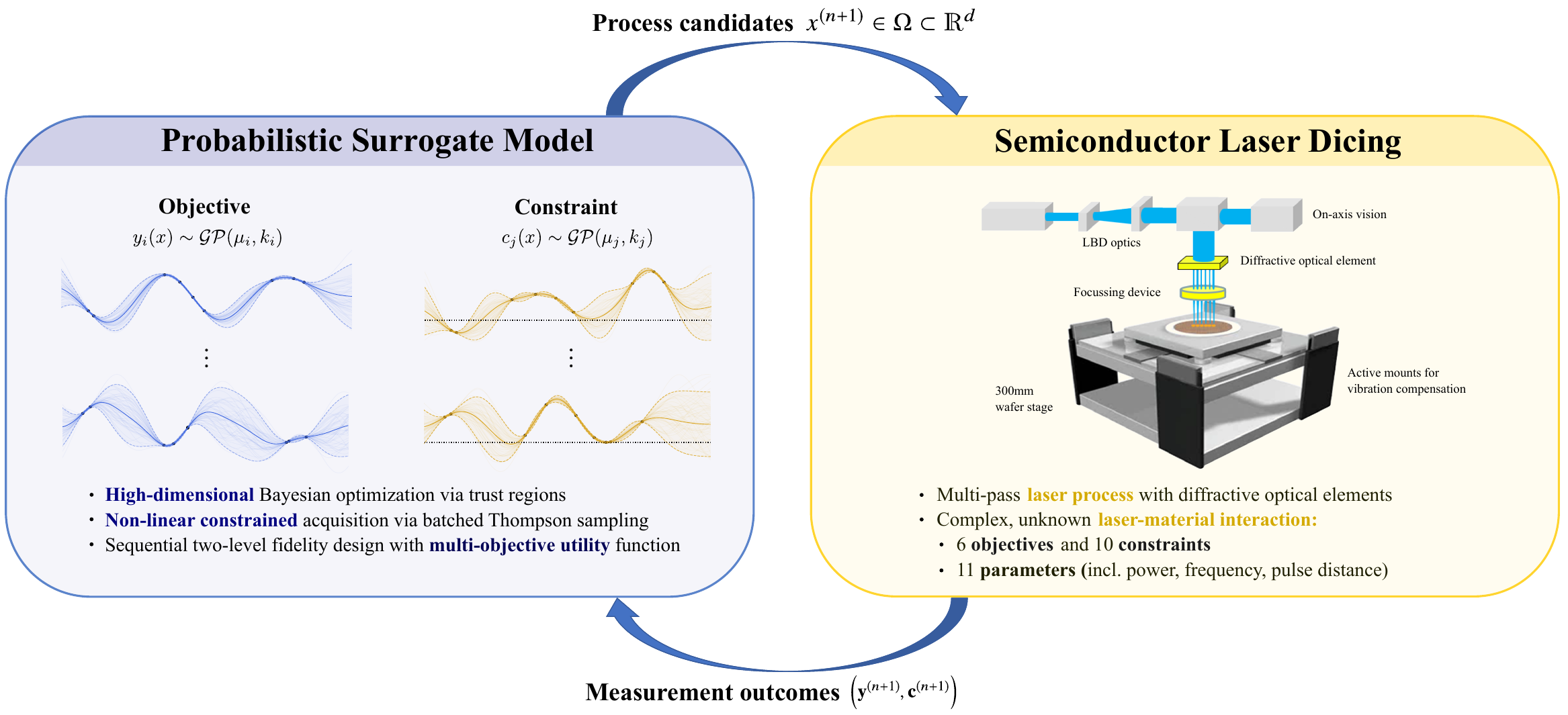}
        \end{tcolorbox}
    \caption{\textbf{Automated discovery of semiconductor laser dicing processes via Bayesian optimization.} Finding optimal process configurations is a critical challenge in semiconductor manufacturing. We propose a closed-loop framework that bridges \textbf{(Left)} Bayesian optimization and \textbf{(Right)} semiconductor laser dicing. A probabilistic surrogate model (Gaussian process) captures the complex laser-material interactions to guide the exploration, while the industrial setup executes the multi-pass dicing process. The goal is to discover a configuration of 11 process parameters (Table~\ref{tab:process_params}) that simultaneously optimizes 6 objectives while strictly adhering to 10 non-linear constraints (Table~\ref{tab:quality_params}). After selecting the most promising candidate, the measurements of the true objective functions and constraint outcomes are fed back into the training of the surrogate model, iteratively refining the search and replacing manual expert tuning with a data-efficient, automated discovery loop.}
        \label{fig1:mainfigure}
  
\end{figure*}

The separation of semiconductor wafers into individual dies is a critical, high-precision step in microelectronic manufacturing~\citep{cao2025laser,yin2024two}. Modern semiconductor wafers consist of up to thousands of integrated circuits (dies) that are fabricated via lithography processes~\citep{moore1965cramming}. Separating these is fundamental to the production of all microchips (i.e., logic chips, memory chips, ASICs and SoCs), sensors, LEDs, and solar cells. This consequently has wide-ranging impact on downstream fields such as medical devices, automotive engineering, consumer electronics, and renewable energy (e.g., photovoltaics)~\citep{she2017review, zeng2017changes, roccaforte2021selective, baliga2023silicon}. 

This separation process is hugely influential on the final qualities of the die, such as its material strength and structural integrity~\citep{marks2014characterization}. It is increasingly performed using multi-pass laser dicing, as this allows for a wide range of control of the separation process~\citep{jain2021laser}. Each laser pass consists of its own set of tunable parameters including power, frequency, and focus, that interact in complex, non-linear ways with the multi-layered wafer materials. Finding a process configuration that balances process speed, material integrity, and separation quality is a significant engineering challenge that involves balancing multiple conflicting objectives. 

While an optimal process configuration can be applied repeatedly once it is established for a specific product, finding such a configuration with desirable quality and throughput for a new material is a major industrial bottleneck~\citep{lei2012singulation}. Often this requires 2-4 weeks of manual expert labor and large amounts of expensive wafer materials to determine a process configuration for a new product~\citep{savriama2015optimization, raghavan2023methodology}. With the trend of increasingly smaller semiconductor wafer technology, the necessity of adapting high-precision laser dicing processes becomes increasingly urgent~\citep{cao2025laser}. Despite this, no general automated solution for laser dicing process discovery exists that is scalable to industrial-scale laser dicing processes.

This work introduces the first automated, data-driven approach to laser dicing process discovery at an industrial scale, using Bayesian optimization (BO)~\citep{shahriari2016takingthehumanout}. This framework defines a Bayesian model (a Gaussian process) and infers the posterior distributions over complex, non-linear laser-material interactions, learning directly from experimental data~\citep{williams2006gaussian}. This data-efficient strategy is crucial for a domain characterized by high evaluation costs (time, material) and the absence of accurate simulators. 
However, successfully applying this BO framework to industrial laser dicing process discovery is not straightforward and requires overcoming four concurrent challenges: (i) The multi-pass laser sequence creates a \textit{high-dimensional} optimization problem -- a known weakness of traditional BO~\citep{malu2021bayesian}. (ii) The process is subject to multiple \textit{non-linear constraints}, including both known physical machine limits and unknown quality constraints that depend on the laser-material interactions, making acceptable solutions difficult to find. (iii) The problem has \textit{conflicting} objectives, such as processing speed, material strength and separation quality. (iv) The quality metrics come from two \textit{different sources}, that is, fast, optical measurements and slow, expensive tests.

Our first main contribution is the Bayesian Optimization for Laser Dicing (BOLD) framework, a novel approach that tackles these four challenges by combining~(a) high-dimensional BO via trust regions;~(b) physical and learned non-linear constraints via Monte Carlo techniques (rejection and Thompson sampling);~(c) multi-objective optimization using an expert-derived weighted utility function;~(d) a sequential two-stage fidelity strategy, leveraging fast optical measurements before incorporating expensive destructive die strength testing. 

The proposed framework is illustrated in Fig.~\ref{fig1:mainfigure}, visualizing the iterative nature of the BO solution.

The second main contribution is an experimental validation on a state-of-the-art industrial laser dicing system (LASER1205). We designed experiments to be representative of industry-scale complexity (a three-pass sequential process) on both bare silicon and complex, multi-layered production wafers -- a task previously unsolved by automated methods. 
The experiments demonstrate that our method autonomously discovers high-quality, feasible laser dicing processes for both materials. On the complex production wafers, the discovered processes match the die strength of the expert-derived baseline, while improving production speed by 34\%. The discovered processes contain qualitatively new insights, contradicting some common engineering beliefs on effective solutions. Furthermore, we show that the discovered processes can be additionally augmented by a human expert, suggesting a synthesis of automatic process control with human expertise to create a solution surpassing either in isolation, in line with the human-in-the-loop machine learning framework~\citep{mosqueira2023human}. Finally, we demonstrate that the framework's final surrogate model can be used to generate a \textit{variety} of process configurations with different trade-offs by altering the utility weights post-hoc, suiting different manufacturing goals.

\section{Related work}
\subsection{Laser dicing optimization in manufacturing} Establishing effective laser dicing process configurations is a well-known industrial challenge, traditionally reliant on expert-driven, iterative manual tuning. Recent works have investigated automation using design-of-experiments (DoE) methodologies. For example,~\cite{raghavan2023methodology} apply this on a two-pass dicing process containing six tunable parameters, while~\cite{savriama2015optimization} construct a similar DoE of four tunable parameters, keeping various other key parameters fixed. Recently,~\cite{yang2024laser} considered a DoE on a LASER1205 machine using the Taguchi method, but considered only three tunable parameters. 

The DoE approach is restricted, however, by the complexity of the process configuration. The limitation of a low number of parameters in prior work is a fundamental weakness of DoE, stemming from its inability to use measurement data to adaptively guide the search. Without this feedback, DoE is forced to sample much more, which due to the curse of dimensionality renders this approach infeasible. For industrial-scale process optimization, where many more tunable parameters are involved than considered in prior works, this renders standard DoE approaches impractical.

In contrast, our proposed BO solution constructs a closed feedback loop, relying on sequential decision-making to efficiently explore this large space. Here, we consider 11 input parameters, which is a significant increase in process complexity compared to prior work. 

\subsection{Bayesian decision-making in manufacturing and control} Data-driven, model-based optimization has lead to new opportunities for accelerating discovery in complex, real-world domains where experiments are expensive. BO in particular has been applied successfully to such problems, including lithography process optimization~\citep{guler2021bayesian}, robotics~\citep{calandra2016bayesian, holzmann2024learning}, cortical neuro-prosthetic vision~\citep{kucucoglu2025}, and the control of large-scale scientific instruments like particle accelerators~\citep{kirschner2019bayesian, jalas2021bayesian, shalloo2020automation}.
More broadly, the core principle of BO -- using a probabilistic model to guide sample-efficient exploration -- is shared with the field of probabilistic model-based reinforcement learning (MBRL), where \textit{dynamic} (time-varying) decisions are optimized, enabling the control of complex, continuous-time problems like robotics~\citep{deisenroth2011pilco, chua2018deep, leeftink2025optimalcontrolprobabilisticdynamics}.

While these applications demonstrate the power of BO and RL, the laser dicing process optimization task presents a unique and concurrent set of challenges that are not solved by standard frameworks, and require a dedicated synthesis of state-of-the-art BO techniques.

\subsection{State-of-the-art Bayesian optimization} Our proposed framework integrates several distinct areas of state-of-the-art BO research to tackle the challenges. First, standard BO struggles with large amount of uncertainty that is indirectly caused by the curse of dimensionality. To account for the high-dimensional problem (11 parameters, described in Table~\ref{tab:process_params}) posed by the sequential multi-pass setup, we build on high-dimensional BO. A commonly adopted approach is sparse-axis aligned subspaces Bayesian optimization (SAASBO)~\citep{eriksson2021high}, which encodes a prior belief that the solution is comprises only a small set of the optimization variables. Alternatively, one can consider trust region-based methods such as TuRBO~\citep{eriksson2019turbo}, which avoids over-exploration by restricting the global search space to a smaller trust region that is placed around the previously best solutions. We adopt this trust-region approach in our work.  

Second, the laser dicing process is governed by non-linear process constraints, including both known physical limits and unknown black-box quality constraints (e.g., avoiding material cracks). This is studied in the field of constrained BO, which learns constraint functions alongside the objective~\citep{gardner2014bayesian}. Our approach is an adaptation of the scalable constraint BO (SCBO) framework~\citep{eriksson2021scbo}, which integrates learned non-linear constraints into the acquisition function. 

Third, the process has multiple conflicting objectives, such as maximizing speed while maintaining die strength. This problem is addressed in the field of multi-objective BO (MOBO), where methods like expected hypervolume improvement (EHVI)~\citep{daulton2020differentiable} efficiently explore the Pareto front, the set of non-dominated solutions where no single objective can be improved without degrading at least one other. Leveraging expert knowledge, we adopt a pragmatic approach by using an expert-derived weighted utility function. 

Finally, the objective and the constraint measurements come from two sources with different costs (described in Table~\ref{tab:quality_params}), forming a multi-fidelity (MF) Bayesian optimization (MF-BO) problem. The field of MF-BO provides principled ways to combine cheap, low-fidelity data with expensive, high-fidelity data~\citep{poloczek2017multi, wu2020practical}. Our two-stage sequential strategy is a practical implementation of this concept, using fast optical measurements to efficiently explore the space before validating promising candidates with expensive die strength tests.

Our work contributes a novel synthesis of various state-of-the-art components. The BOLD framework is, to our knowledge, the first to successfully integrate these techniques to solve the complex, high-dimensional, and constrained industrial-scale problem of automated laser dicing process discovery.


\section{Laser Dicing of Semiconductor Wafers}\label{sec:laserseparation-preliminaries}
\begin{figure}[t]
    \centering
     \begin{subfigure}[b]{0.625\linewidth}
         \centering
         \includegraphics[width=\textwidth]{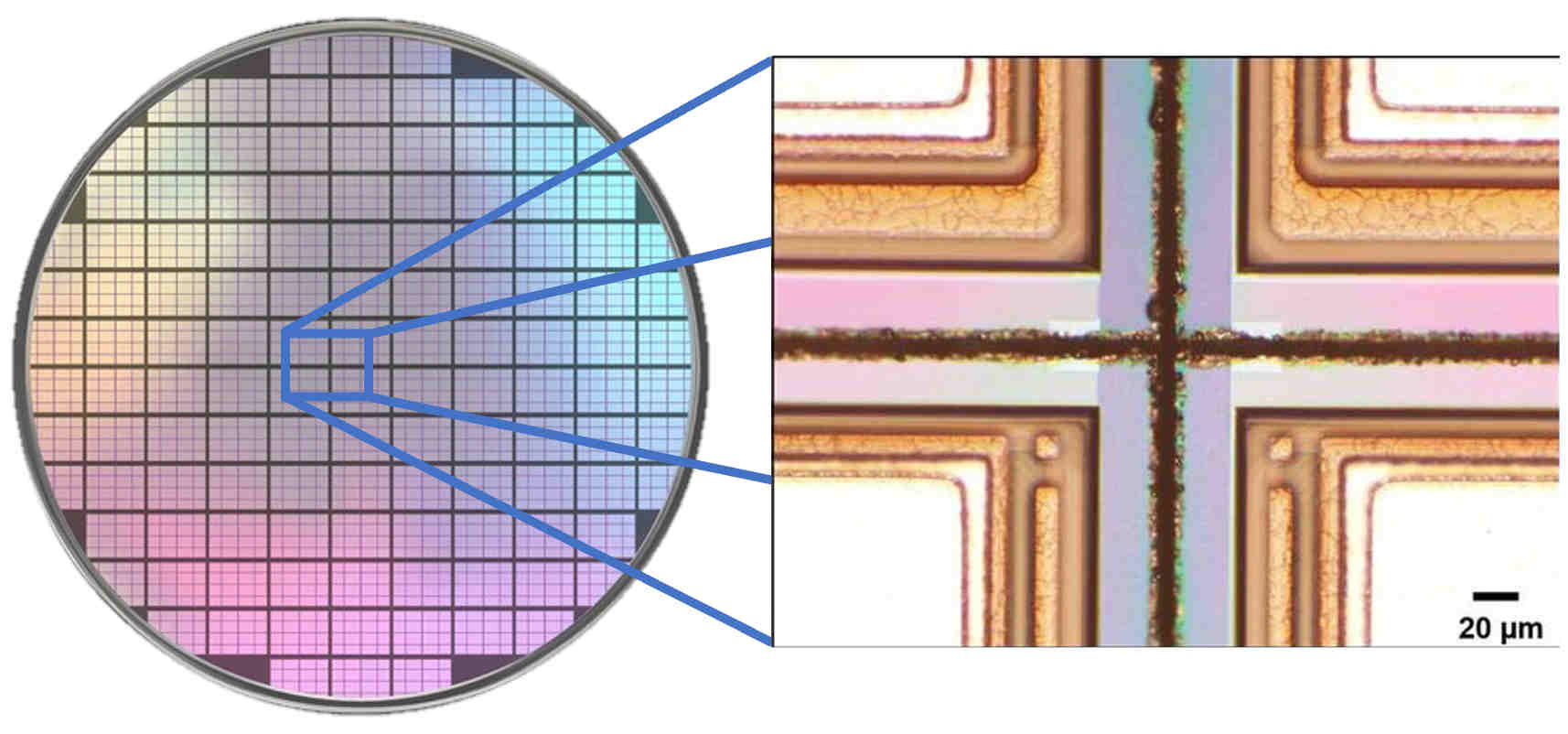}
         \label{fig:2a}
     \end{subfigure}
     \hfill
     \begin{subfigure}{0.3\linewidth}
         \centering
         \raisebox{-4pt}{\includegraphics[width=\textwidth]{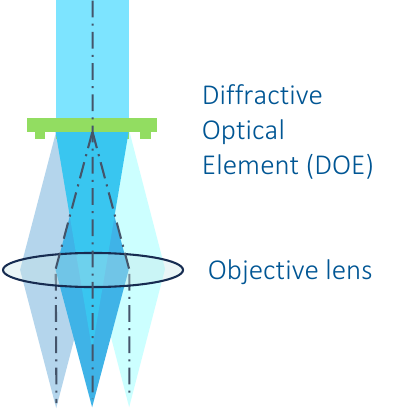}}
         \label{fig:2b}
     \end{subfigure}
    \caption{\textbf{(Left)} Separated dies on a semiconductor wafer, and post-processed die using laser dicing. \textbf{(Right)} Diffractive optical element (DOE) and objective lens used for beam-splitting, a process that maximizes available laser power while minimizing the heat-affected zone.}
    \label{fig:2laserdicing-and-vdoe}
\end{figure}

Laser dicing of processed semiconductor wafers is a critical step in microelectronic manufacturing, which aims to precisely separate integrated circuits from a wafer~\citep{cheng2013review}. Wafers consist of up to thousands of integrated circuits (dies) fabricated via lithography processes, which leave the dies organized in squares separated by dicing streets. After separation, the dies can be used in individual microchips, sensors or LEDs.

Modern wafers are complex, heterogeneous and multi-layered structures, which typically consist of a bulk substrate (e.g., silicon), a stack of dielectric layers on top, and possibly a metal layer on the back. This intricate multi-layered circuit design significantly complicates the separation process, since the laser-material interaction involves a variety of complex, physical processes depending on the laser regime. Examples of these processes include avalanche ionization~\citep{yao2005time, wright2019memory}, multiphoton ionization~\citep{klaiber2016crossover}, Coulomb explosion~\citep{wang2021evolution}, and dielectric breakdown~\citep{tang2021stochastic}. These processes ultimately determine the final quality of the separated die.

These laser dicing challenges can be addressed by applying multiple sequential laser passes to achieve a clean and robust separation without damaging the circuitry~\citep{jbor2015multi}.
To achieve the required precision during the dicing process, the laser optics remain stationary, while the wafer is moved underneath on a high-precision motion-control stage. As a result, the main challenge of laser dicing is not one of physical positioning -- which can be solved using high-precision motion control -- but in establishing the sequential laser pass process parameters that effectively and robustly separate the die on a given material. 

\subsection{Multi-pass laser dicing}
A single laser pass employs nanosecond short pulses and highly concentrated energy, enabling high-precision material processing while minimizing thermal damage like cracks and melting. A sequence of multiple laser passes can be used to effectively handle the intricate material composition. In its simplest form, this process can be broken down into three successive passes:
\begin{enumerate}
    \item \textbf{Trenching:} The initial pass aims to create a shallow kerf in the dielectric top layers on both sides of the final cut. This step clears a path and pre-defines the separation line, minimizing damage to the top surface in subsequent passes.
    \item \textbf{Dicing:} The second pass separates the wafer and removes the bulk material between the trenches and, if existing, the backside metal.
    \item \textbf{Recovery:} The final pass is used to recover damage from the top side of the dicing kerf, by evaporation and remelting the edge of the kerf \citep{jbor2015multi, jbor2017vdoe}.
\end{enumerate}
While laser processes with more than three passes can be constructed as well depending on the wafer material, the additional passes fall into the categories of trenching, dicing, or recovery passes.
Crucially, the three steps are not independent, but maintain complex dependencies between one another. In order to adapt a laser dicing process to a new wafer material, the solution to a complex, high-dimensional optimization problem has to be found, where all the adjustable parameters of each laser pass (e.g., laser power, frequency, focus position) have to be jointly co-adapted. 

The complex interplay between passes is necessary to manage the intense laser-material interactions. The main dicing pass, for example, causes material to melt, creating a heat-affected zone (HAZ) and recast -- molten material that resolidifies on the die edge. This area is generally associated with reduced die strength. Consequently, process configurations with excessive energy can easily degrade die quality. 

To effectively manage these high pulse energies and mitigate the formation of HAZ, modern laser dicing systems leverage diffractive optical elements (DOEs)~\citep{jbor2015multi}. A DOE is a part of the optical path which shapes the laser beam by splitting into a smaller, pre-defined pattern, which allows the system to effectively distribute the available pulse energy as depicted in Fig.~\ref{fig:2laserdicing-and-vdoe}. This technology is a key enabler for the multi-pass approach, as each pass (trenching, dicing, recovery) typically uses a specific DOE pattern, designed and chosen to match each other. The recovery pass, for instance, serves as a key element in ensuring high die strength by specifically treating the HAZ. It often uses a particularly effective pattern known as the V-DOE. This pattern, named after its two-dimensional shape, ensures that the complete edge of the dicing kerf is treated to maximize recovery~\citep{jbor2015multi}. 


\begin{figure}[t]
    \centering
    \includegraphics[width=\linewidth]{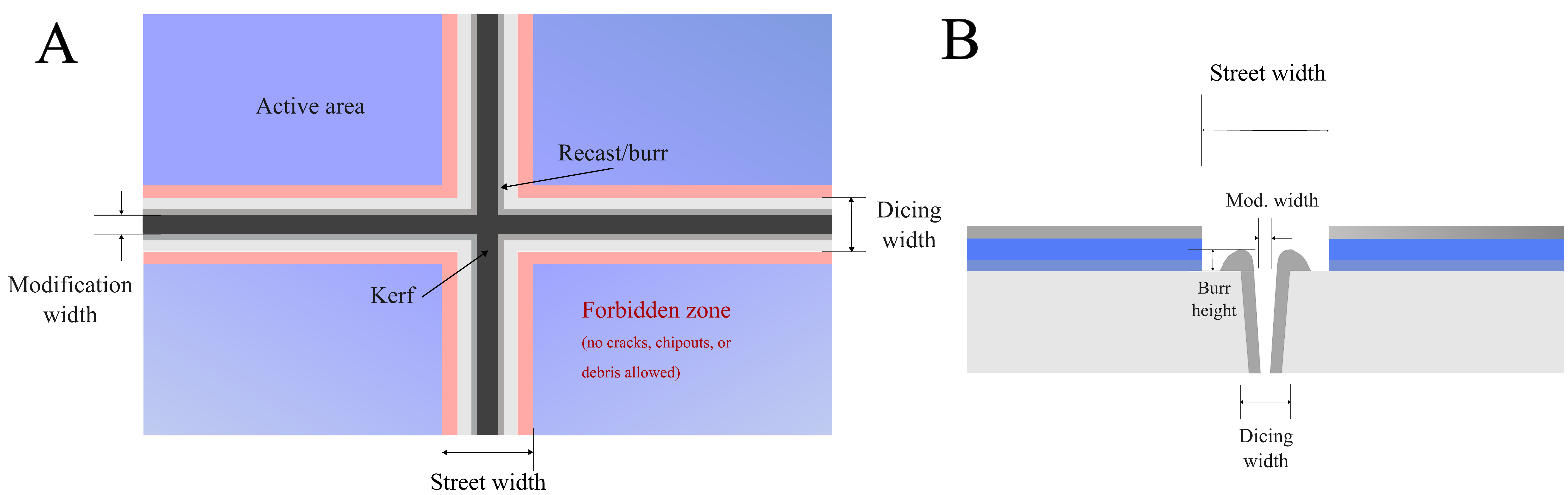}
    \caption{\textbf{Laser dicing streets and key metrics}. Top-down \textbf{(left)} and  cross-section \textbf{(right)} views of the laser dicing street. The two perspectives illustrate the key metrics (dicing, width, kerf width, burr height) and structural defects (cracks) that define the quality of the dicing process.}
    \label{fig:micropscopic-measurements}
\end{figure}
\subsection{Measuring process quality} \label{sec:3.2processquality}
The quality of a given laser dicing process is evaluated across two distinct types of measurement: optical inspection for geometric and structural defects, and destructive die strength testing for mechanical integrity. The third key metric, throughput (production speed) can be computed analytically as a function of the process parameters, denoted as $t(\mathbf{x})$.

Optical inspection assesses the dicing streets for both geometric quality and critical structural defects. Geometric quality is evaluated using continuous values, including the dicing width (the width of the primary laser cut), the modification width (the total area visibly affected by the laser), and the recast/burr (the height of molten material that resolidified on the die edge). 
Structural defects are also measured, such as chipouts (where portions of the top-layer material are chipped away near the dicing street) and cracks propagating into the active die area. These defects, which can occur on the inside, corner, or backside of the die, are crucial to measure as they are caused by different mechanisms during the laser dicing process. A visual depiction of these quantities is shown in Fig.~\ref{fig:micropscopic-measurements}.

Arguably the primary measure of a die's mechanical integrity for production quality is die strength~\citep{domke2016ultrashort}. This destructive test involves applying a precise force (measured in Newtons) to a sample of separated dies using a three-point bending test until they fracture. This breaking force is then used to calculate the die's flexural strength, which is typically reported in megapascals (MPa). It is crucial to measure both the front-side (top) and back-side (bottom) die strengths, as they are affected by different failure modes; a given process configuration might produce high front-side strength but low back-side strength, or vice versa.


\section{Bayesian Optimization}\label{sec:bo-preliminaries}
Bayesian optimization (BO) is a sequential, model-based strategy for finding the optimum of an expensive-to-evaluate black-box function, $f(\mathbf{x})$, within a bounded domain \mbox{$\Omega \subset \R^d$}~\citep{shahriari2016takingthehumanout}. Its data-efficiency comes from intelligently selecting the next input to evaluate , that is, the next set of process parameters, by balancing the trade-off between \textit{exploration}, querying uncertain regions of the parameter space, and \textit{exploitation}, querying regions likely to contain the optimum. This is also referred to as dual control in control theory literature~\citep{Wittenmark1995Adaptive}.

Instead of relying solely on local gradient information, BO constructs and iteratively refines a global probabilistic surrogate model of the objective function. This process can be broken down into two components: 
\begin{enumerate}
    \item \textbf{Probabilistic Surrogate Model}: A statistical model, typically a Gaussian process (GP), is fitted to all available data points. This surrogate approximates the true objective function and provides uncertainty estimates for its predictions.

    \item \textbf{Acquisition Function:} This function, $\alpha(\mathbf{x}; \mathcal{D}_n)$, uses the surrogate model's predictions and uncertainty to quantify the utility of evaluating any candidate point. It is designed to not simply choose the best function value observed so far, but to explicitly balance the trade-off between exploiting with exploration. The next point to query is chosen by maximizing this function.
    
\end{enumerate}
This iterative process of updating the probabilistic surrogate model with new data and maximizing the acquisition function to select the next query forms the core BO loop, as illustrated in Fig.~\ref{fig1:mainfigure}. 

\subsection{Gaussian processes}
Gaussian processes are used as the surrogate model in BO due to their flexibility, data-efficiency, and ability to quantify uncertainty~\citep{williams2006gaussian}. A GP defines a prior distribution over the space of functions directly, and is written as:
\begin{equation}
    f(\mathbf{x}) \sim \mathcal{GP}\left(\mu_0 (\mathbf{x}), k(\mathbf{x}, \mathbf{x'})\right),
\end{equation}
where $\mu_0 (\mathbf{x})$ is the prior mean function (typically zero) and $k(\mathbf{x}, \mathbf{x}')$ is the covariance function or kernel. The kernel encodes the relationship between different outputs as a function of the covariance of their corresponding inputs. It thereby expresses a prior belief about the function's structural properties, such as smoothness or periodicity~\citep{williams2006gaussian}. A common choice in BO is the Matèrn 5/2 kernel:
\begin{align}
    k_{\text{Matèrn 5/2}}(\mathbf{x}, \mathbf{x}') &= \sigma^2_f \left( 1 + \sqrt{5} r + \frac{5}{3} r^2 \right) \exp \left( -\sqrt{5} r \right) ,  \label{eq:matern52_kernel}
\end{align}
where $r^2 = \sum^d_{j=1} \flatfrac{(x_j - x'_j)^2}{\ell_j^2} $ is the squared distance between inputs scaled by a lengthscale $\ell_j$ unique for each input dimension, and the signal variance $\sigma_f^2$ determines the total variance of the function. This kernel encodes the belief that if two input values $\mathbf{x}$ and $\mathbf{x}'$ are similar, their output values will also correlate -- effectively imposing a functional prior that the modeled function is smooth.

In the cycle of Bayesian belief updating, the prior is updated into a posterior distribution as new data is observed. Given a dataset $\mathcal{D}^{(n)} =\{(\mathbf{x}^{(i)}, y^{(i)})\}^n_{i=1},$ where each $\mathbf{x}_i$ represents a laser process configuration and $y_i$ is a measured quality score, the goal is to condition the GP on these observations. Assuming a Gaussian likelihood function for the (possibly noisy) measurements, this yields a closed-form posterior predictive distribution for any test input $\mathbf{x}$, with mean and covariance: 
\begin{align}
    \mu_n (\mathbf{x}) &= \mu_0 (\mathbf{x}) +   \mathbf{k}(\mathbf{x})^{\top} \left(\mathbf{K}+\sigma^2 \mathbf{I}\right)^{-1}  (\mathbf{y}-\mathbf{m}) , \label{eq:mean_gp}\\  
    \sigma^2_n (\mathbf{x}) &= k(\mathbf{x},\mathbf{x}) - \mathbf{k}(\mathbf{x})^{\top}  \left(\mathbf{K} +\sigma^2 \mathbf{I}\right)^{-1}  \mathbf{k}(\mathbf{x}),  \label{eq:cov_gp}
\end{align}
where $\mathbf{K}$ is the $n \times n$ covariance matrix of the observed inputs with elements $K_{i,j} \coloneqq k(\mathbf{x}_i, \mathbf{x}_j)$, $\mathbf{k}(\mathbf{x}) = [k(\mathbf{x},\mathbf{x}_1),\dots, k(\mathbf{x},\mathbf{x}_n)]$ is the $n \times 1$ vector of covariances between the new input $\mathbf{x}$ and each observed input $\mathbf{x}_i$, $\mathbf{y}$ is the vector of observed outcomes, and $m_i \coloneqq \mu_0 (\mathbf{x}_i)$ are the evaluations of the prior mean function at each observed input. 
The posterior mean $\mu_n (\mathbf{x})$ represents the model's best prediction of a configuration's outcome, while the variance $\sigma^2_n (\mathbf{x})$ quantifies the model's uncertainty in that prediction. This uncertainty estimate is critical for guiding the BO process, as it allows the acquisition function to make informed decisions that effectively balance exploration and exploitation. 

The posterior distribution depends on the kernel's hyperparameters $\theta$, such as the variance $\sigma^2_f$ and the lengthscales $\ell_j$. These values significantly influence the model fit on the data; for example, a large value for the lengthscale induces a slowly changing function, whereas a low value induces a fast changing function. These hyperparameters are not set to a fixed value but can be optimized using the log marginal likelihood for the inputs $\mathbf{X}\coloneqq \left\{ \mathbf{x}_i\right\}^n_{i=1}$: 
\begin{equation}
    \log p(\mathbf{y}\mid\mathbf{X}, \theta) \propto  - \overbrace{\frac{1}{2} \mathbf{y}^T \left(\mathbf{K}+\sigma^2_n \mathbf{I}\right)^{-1} \mathbf{y}}^\text{Model fit} - \overbrace{\frac{1}{2} \log \left|\mathbf{K}+\sigma^2_n \mathbf{I}\right|}^\text{Complexity penalty} , \label{eq:loglikelihood}
\end{equation}
where $\sigma^2_n$ is the observation noise. This optimization, known as type II maximum likelihood~\citep{mackay1999comparison}, is typically performed using gradient-based methods. The hyperparameters are iteratively updated until convergence, ensuring that the GP model best explains the observed data. 

\subsection{Acquisition function}
The acquisition function uses the probabilistic surrogate model to select the next point to evaluate. Instead of relying on local gradient information, it leverages the GP's global predictions and uncertainty estimates to guide the search efficiently. Various acquisition functions have been developed, each implementing a different heuristic for balancing the exploration-exploitation trade-off. 

Well-known acquisition examples include expected improvement (EI)~\citep{movckus1974bayesian}, probability of improvement (PI)~\citep{kushner1964new}, entropy search (ES)~\citep{hennig2012entropysearch} and predictive entropy search (PES)~\citep{hernandez2014predictive}, knowledge gradient (KG)~\citep{frazier2009knowledge}, and upper confidence bound (GP-UCB)~\citep{auer2002using,srinivas2012ucb}. The latter is valued for its theoretical connections to bandit literature and its intuitive form, which explicitly sums exploitation and exploration terms:
\begin{equation}
    \alpha_{\text{UCB}}^{(n+1)}(\mathbf{x}) \coloneqq \mu^{(n)} (\mathbf{x}) +  \sqrt{\beta^{(n+1)}} \sigma^{(n)} (\mathbf{x}),
\end{equation}
where $\beta^{(n)} \geq 0$ is a manually chosen hyperparameter at iteration $n$. 
A different acquisition strategy that is of particular importance to this work is Thompson sampling (TS)~\citep{thompson1933likelihood}, which leverages Bayesian posterior sampling:
\begin{equation}
    \mathbf{x}^{(n+1)} = \argmax_{\mathbf{x}\in \Omega} \tilde{f}^{(n)} (\mathbf{x})
    \text{ where }\tilde{f}^{(n)}\sim \mathcal{GP}\left(\mu_0, k \mid \mathcal{D}^{(n)}\right),
\end{equation}
where $\tilde{f}^{(n)}$ is a sampled function from the GP posterior after $n$ iterations. This implicitly balances exploration and exploitation by drawing a sample from the posterior distribution over optimal input locations. In this work, we leverage Thompson sampling due to its ability to incorporate non-linear constraint functions.

\section{Bayesian Optimization for Laser Dicing}\label{sec:4mainmethod}
\subsection{Problem formulation and modeling approach}\label{sec:4.1subsection}
The automated discovery of laser dicing processes is formulated as a high-dimensional, constrained, multi-objective black-box optimization problem. The core task is to find a set of process parameters, $\mathbf{x} \in \Omega \subset \R^{d}$, that maximizes a set of competing objectives, such as production speed, die strength, and separation quality. The exact relationships between these parameters and the process outcomes are unknown, and are governed by complex, non-linear laser-material interactions that have to be learned empirically.

Our framework addresses this by maintaining two distinct datasets. First, the objective dataset aggregates all information related to the \textit{unknown} performance goals, namely the dicing width, modification width, burr/recast height, and front and backside die strength (summarized in the top part of Table~\ref{tab:quality_params}). Formally, we write this as $\mathcal{D}^{(n)}_{\text{O}} \coloneqq \left\{\left(\mathbf{x}^{(i)}, \mathbf{y}_{\text{O}}^{(i)}\right)\right\}^n_{i=1}$, where $$\mathbf{y}_{\text{O}}^{(i)} = \left(y^{(i)}_{\text{dicing-width}}, y^{(i)}_{\text{mod-width}}, y^{(i)}_{\text{burr}}, y^{(i)}_{\text{front-strength}}, y^{(i)}_{\text{back-strength}}\right)$$ for $i = 1, \dots, n$. Since the outputs are independent physical properties, we adopt a multi-output GP with independent posteriors
\begin{equation}
    \mathbf{f} \sim \mathcal{GP}_{\text{O}}\left(\mu_0 (\mathbf{x}), k(\mathbf{x}, \mathbf{x'}) \mid \mathcal{D}_{\text{O}}\right), \label{eq:gp_objectives}
\end{equation}
where $\mathbf{f} \colon \R^d \to \R^{d_O}$ is defined by $\mathbf{f}(\mathbf{x}) \coloneqq \left(f_1(\mathbf{x}), \dots, f_{d_O} (\mathbf{x})\right)^{\top} $, and $d_{\text{O}}=5$ is the number of unknown objective functions. Second, the constraint dataset aggregates all information related to the \textit{unknown} constraint violations, namely front side material cracks, front side corner cracks, back side material cracks, backside separation, and restricted chipouts (summarized in Table~\ref{tab:quality_params}, bottom part). Constraints differ from objectives by modeling the violation of requirements that \textit{have to be} satisfied, rather than optimized. Input configurations which satisfy every constraint are known as \textit{feasible} configurations, defining acceptable product requirements. We formalize this as $\mathcal{D}_{\text{C}} \coloneqq \left\{\left(\mathbf{x}^{(i)}, \mathbf{y}^{(i)}_{\text{C}}\right)\right\}_{i=1}^{n}$, where $$\mathbf{y}_{\text{C}}^{(i)} = \left(y^{(i)}_{\text{front-cracks}}, y^{(i)}_{\text{corner-cracks}}, y^{(i)}_{\text{back-cracks}}, y^{(i)}_{\text{separation}}, y^{(i)}_{\text{chipouts}}\right)$$ for $i = 1, \dots, n$. Each target $c$ encodes the violation of the failure mode, i.e., the difference between the permitted violation and the observed violation, keeping the targets as a continuous regression. We again adopt a multi-output GP with independent posteriors 
\begin{equation}
    \mathbf{c} \sim \mathcal{GP}_{\text{C}}\left(\mu_0 (\mathbf{x}), k(\mathbf{x}, \mathbf{x'}) \mid \mathcal{D}_{\text{C}}\right),
\end{equation}
where $\mathbf{c} \colon \R^d \to \R^{d_c}$ is defined by $\mathbf{c}(\mathbf{x}) \coloneqq \left(c_1(\mathbf{x}), \dots, c_{d_C} (\mathbf{x})\right)^{\top} $, and $d_{\text{C}} = 5$ is the number of unknown constraint functions.

By adopting independent posteriors in the multi-output GP models, one can alternatively interpret both models as collections of independent GPs. This multi-output approach with independent posteriors is essential for learning objective-specific properties, such a smoothness and noise level. This also naturally accommodates the different data set sizes inherent to the multi-fidelity strategy described later in this section.

\begin{algorithm*}[t]
    \begin{small}
    \caption{Bayesian Optimization for Laser Dicing (BOLD)}
    \label{alg:laser_bo}
    \begin{algorithmic}[1]
        \Require 
            $\Omega$ (parameter space),
            $n_{\text{init}}$ (number of initial points), 
            $\mathcal{T}_{\text{threshold}}$,
            $u(\cdot ; \mathbf{w})$ (expert-weighted utility function),
            $t(\cdot)$ (analytical throughput function), 
            $q$ (batch size)
        \State Generate initial configurations $\left(\mathbf{x}^{(1)},  \ldots , \mathbf{x}^{(n_{\text{init}})}\right)$ using a Sobol sequence or database configurations
        \State Evaluate $\mathbf{x}^{(i)}$ on physical laser system and observe $\mathbf{y}_{\text{optic}}^{(i)}$, $\mathbf{c}_{\text{optic}}^{(i)}$ for $i \in [1, \ldots, n_{init}]$
        \State $\mathcal{D}^{(n_{init})}_{\text{O}} \gets  \bigcup_{i=1}^{n_{init}} \left\{\left(\mathbf{x}^{(i)}, \left(\mathbf{y}_{\text{optic}}^{(i)}, \emptyset \right) \right) \right\}; \enspace \mathcal{D}^{(n_{init})}_{\text{C}} \gets  \bigcup_{i=1}^{n_{init}} \left\{\left(\mathbf{x}^{(i)}, \left(\mathbf{c}_{\text{optic}}^{(i)}, \emptyset \right) \right) \right\}$ \Comment{Initial data sets}
        \State Initialize trust region side length $\tau^{(n_{\text{init}})}$, that defines the trust region hypercube $\mathcal{T}^{(n_{\text{init}})} \subseteq \Omega$ 
        \State $n \gets n_{\text{init}}$ 
        \While{$\tau^{(n)} \geq \tau_{\text{threshold}}$ and sample budget remaining}
            \State Fit objective model $\mathcal{GP}_{\text{O}}\left(\mu_0, k \mid \mathcal{D}^{(n)}_{\text{O}}\right)$ and constraint model $\mathcal{GP}_C{}\left(\mu_0, k \mid \mathcal{D}^{(n)}_{C}\right)$ 
            \State Initialize hypercube trust region $\mathcal{T}^{(n)}$ centered on the best-so-far feasible evaluation
            \State Generate $r$ candidate points $\tilde{\mathbf{x}}^{\text{cand}} \gets \{\tilde{\mathbf{x}}_1, \dots, \tilde{\mathbf{x}}_r\} $ in the trust region $\mathcal{T}^{(n)}$
            \State Filter on \textit{known} non-linear constraints to form feasible set $\mathcal{F} \gets \left\{\tilde{\mathbf{x}} \in \tilde{\mathbf{x}}^{\text{cand}} \mid \mathbf{c}_{\text{known}}(\mathbf{\tilde{x}}) \leq \mathbf{0}\right\}$  

            \For{$j = 1 \to q$} \Comment{Batch constrained Thompson sampling}
                \State Draw surrogate objective sample $\tilde{\mathbf{f}}^{(j)} \sim \mathcal{GP}_{\text{O}}\left(\cdot \mid \mathcal{D}_{\text{O}}^{(n)}\right)$ and constraint sample $\tilde{\mathbf{c}}^{(j)} \sim \mathcal{GP}_{C}\left(\cdot \mid \mathcal{D}_{C}^{(n)}\right)$
                \State Filter on \textit{learned} constraint: $\mathcal{F}_{\text{sample}}^{(j)} \gets \left\{\tilde{\mathbf{x}} \in \mathcal{F} \mid \tilde{\mathbf{c}}^{(j)}(\tilde{\mathbf{x}}) \leq \mathbf{0} \right\}$
                \State Select best point from sample: $\mathbf{x}^{(n+j)} \gets \argmax_{\mathbf{x} \in \mathcal{F}_{\text{sample}}^{(j)}} u\left(\tilde{\mathbf{f}}^{(j)}, \mathbf{x}, t; \mathbf{w}\right)$ \Comment{Weighted multi-objective utility}
            \EndFor

            \State Evaluate batch configurations $\left\{\mathbf{x}^{(n+j)}\right\}_{j=1}^q$ on physical laser system \Comment{Physical experiment}
            \State Observe optical data $\left\{\mathbf{y}_{\text{optic}}^{(n+j)}, \mathbf{c}_{\text{optic}}^{(n+j)}\right\}_{j=1}^q$ for the batch
            \State $\left(\left\{\mathbf{y}_{\text{destr.}}^{(n+j)}\right\}_{j=1}^q, \left\{\mathbf{c}_{\text{destr.}}^{(n+j)}\right\}_{j=1}^q \right)  \gets 
                \begin{cases}
                    (\{\emptyset\}, \{\emptyset\}) & \text{if Stage~1}   \\
                    \text{Observe destructive data for $\mathbf{x}^{(n+j)}$} & \text{otherwise } 
                \end{cases}$ 
            \State $\mathcal{D}^{(n+q)}_{\text{O}}  \gets \mathcal{D}^{(n)}_{\text{O}} \cup \bigcup_{j=1}^{q} \left\{\left(\mathbf{x}^{(n+j)}, \left(\mathbf{y}_{\text{optic}}^{(n+j)}, \mathbf{y}_{\text{destruct.}}^{(n+j)}\right)\right)\right\}; \enspace \mathcal{D}^{(n+q)}_{C} \gets \mathcal{D}^{(n)}_{C} \cup \bigcup_{j=1}^{q} \left\{\left(\mathbf{x}^{(n+j)}, \left(\mathbf{c}_{\text{optic}}^{(n+j)}, \mathbf{c}_{\text{destruct.}}^{(n+j)}\right) \right)\right\}$    
            \State Update trust region side length $\tau^{(n+q)}$ 
        \EndWhile
    \end{algorithmic}
    \end{small}
\end{algorithm*}
The following subsections detail the components of our proposed solution, which combines high-dimensional BO, multi-objective scalarization, constrained acquisition functions, and a two-stage fidelity strategy.
The Bayesian optimization of laser dicing (BOLD) method is summarized in Algorithm~\ref{alg:laser_bo}.

\subsubsection{High-dimensional BO via trust regions}\label{sec:4.3subsection}
The application of BO to black-box functions with high-dimensional search spaces is challenging. To address these challenges, trust region BO (TuRBO) combines principles of local optimization with the global exploration capabilities of BO \citep{eriksson2019turbo}. This method restricts the search space to a confined region, the trust region $\mathcal{T}\subseteq \Omega$, which dynamically moves around the objective landscape.
The trust region is centered around the best evaluation found thus far and adapts dynamically. On successive improvements, the region of interest increases, whereas it shrinks if the optimal solution thus far has not improved in recent iterations. A common choice for the trust region is a hypercube, although recent work has explored alternatives in the form of a (truncated) normal distribution \citep{rashidi2024cylindrical}.

Here, we adopt the trust region approach to scale the optimization to a problem with $d=11$ input parameters without overexploring, representing the tunable laser dicing process parameters. This dimensionality is on the border of the number of input dimensions that can effectively be handled by traditional BO. However, even in this parameter range, TuRBO has shown to attain improved data-efficiency \citep{santoni2024comparison}, enabling efficient exploration over the separation process parameters.

\subsubsection{Scalarization of competing objectives}
The laser dicing process is characterized by multiple, conflicting objectives, such as maximizing production speed while simultaneously maximizing die strength and maintaining product integrity. To handle these competing objectives, our framework employs an expert-weighted utility function. 

This approach combines and summarizes the known throughput objective function $t(\cdot)$ (see Section~\ref{sec:3.2processquality}) as well as the learned unknown objectives captured by $\mathbf{f}$ (see Eq.~\eqref{eq:gp_objectives}). The weighted utility is implemented as:
\begin{align} \label{eq:weighted_sum}
    u(\mathbf{f}, \mathbf{x}, t ; \mathbf{w}) &= w_0 t(\mathbf{x}) + \sum_{i=1}^{M} w_i \mathbf{f}_i(\mathbf{x}).
\end{align}
The weights $(w_0, \dots, w_M)$ are pre-determined and derived by process experts by iteratively calibrating the calculated utility scores against their domain assessment of quality for a range of hypothetical measurement outcomes (described in Table~\ref{tab:utility-weights}). This integrates the various objectives (i.e., optical objectives, throughput objectives and destructive objectives) into a single function, which can be leveraged for optimal candidate selection to generate new solution candidates. Crucially, this still allows the model to learn independent posteriors over the various physical processes, such as dicing width or die strength. As such, this method separates the data-driven learning of physical properties from the expert-defined engineering goals.

\subsubsection{Constraints via rejection and Thompson sampling}\label{sec:4.4subsection}
The optimization framework must operate within the physical limits of the laser system, defined as a known set of non-linear inequality constraints. These are derived a priori from fundamental machine properties and are essential for ensuring operational safety. For process discovery, we are interested in estimating the feasible set $\mathcal{F}_{\text{known}} = \left\{\mathbf{x} \in  \mathcal{T}^{(n)} \mid c_{\text{known}}(\mathbf{x}) \leq \mathbf{0}\right\}$. For the machine considered in this work, there are five such proprietary non-linear physical constraints. 

We incorporate these constraints using rejection sampling, a Monte Carlo sampling-based approach. Within the trust region $\mathcal{T}$, we generate a large set of $r$ candidate points $\tilde{\mathbf{x}}^{\text{cand}}$ (e.g., via a Sobol sequence or uniform sampling) within the trust region $\mathcal{T}^{(n)}$. This set is then filtered to form the feasible set, which is then used by the acquisition function (see lines 8--9 in Algorithm~\ref{alg:laser_bo}).

In addition to the known physical constraints, the optimization process is subject to a second set of non-linear inequality constraints. These arise from the complex laser-material interaction and are not known a priori, and include critical failure modes such as material cracks, chipouts, or incomplete backside separation. These constraints must be learned from data. 
To this end, a multi-output GP with independent posteriors $\mathcal{GP}_C \left(\mu_0, k \mid \mathcal{D}^{n}_C\right)$ is used as a surrogate model of the constraint violations (depicted in orange in Fig.~\ref{fig1:mainfigure}). This approach treats constraint violation as a regression problem, where negative output values represent feasible configurations and positive values indicate a violation.

To include the learned non-linear inequality constraints in candidate selection, we employ constrained Thompson sampling. Following the SCBO framework, we augment the Thompson sampling procedure by drawing posterior samples from both the posterior distribution over the objective model and the constraint model~\citep{eriksson2021scbo}. The next evaluation is selected by optimizing the sampled objective function, subject to the sampled constraint function being satisfied. This is formulated as:
\begin{align}
    \mathbf{x}&^{(n+1)} = \argmax_{\mathbf{x}\in \mathcal{F}} u\left(\tilde{\mathbf{f}}_{\text{O}}^{(n)}, \mathbf{x}; \mathbf{w}\right) && \text{where }\tilde{\mathbf{f}}_{\text{O}}^{(n)}\sim \mathcal{GP}\left(\mu_0, k \mid \mathcal{D}^{(n)}_{\text{O}}\right), \nonumber\\ 
    &\text{subject to} \enspace \tilde{\mathbf{c}}^{(n)}(\mathbf{x}) \leq \mathbf{0} && \text{where } \tilde{\mathbf{c}}^{(n)}\sim \mathcal{GP}\left(\mu_0, k \mid \mathcal{D}^{(n)}_{C}\right), 
\end{align}
where $\tilde{\mathbf{f}}^{(n)}$ represents the vector of \textit{sampled} objective functions from the GP posterior $\mathcal{GP}_{\text{O}}\left(\cdot \mid \mathcal{D}_{\text{O}}\right)$, and $\tilde{\mathbf{c}}^{(n)}$ represents the vector of \textit{sampled} constraint functions from the GP posterior $\mathcal{GP}_{C}\left(\cdot \mid \mathcal{D}_{C}\right)$. The optimization is performed over the set $\mathcal{F}$, ensuring the candidates comply with physical machine limitations. 

For laser dicing process optimization, we apply the GP formulation to the backside separation, the front side material cracks, the front side corner cracks, the backside material cracks, and chipouts, each modeled as a function of all input variables. Furthermore, to decrease the total experiment duration, we parallelize this procedure using batched Thompson sampling. This selects $q$ process configurations simultaneously, by drawing $q$ samples from the posterior distributions and independently evaluating their optima.

\subsubsection{Two-stage fidelity strategy for laser dicing processes}\label{sec:4.5subsection}
The laser dicing process is characterized by two information sources with different evaluation cost. Optical measurements, including all learned constraint functions (for example, cracks, chipouts) and some geometric objectives (for example, dicing width), are relatively fast and material-efficient. In contrast, the destructive die strength tests are costly, with a single high-fidelity evaluation requiring ten or more dies to be individually prepared and tested on a three-point bending apparatus, a procedure repeated for both front and back sides. This forms a significant time and material bottleneck.

Evaluating a process configuration with these destructive tests is particularly wasteful when the process configuration fails to meet the optical constraints. To address this, we propose a sequential two-stage fidelity approach where the optimization is split into two stages. In the initial stage, the framework optimizes an initial surrogate objective formulated using \textit{only} optical measurements (Stage~1). This initial stage allows the algorithm to learn the feasible region and identify promising areas using only cheap-to-acquire data.

Once the trust region has shrunk below a predefined threshold, the surrogate objective is replaced with the full objective function containing both optical and destructive measurements (Stage~2). This approach significantly reduces the required number of die strength tests, since the multi-output GP model can condition on all the optical data from Stage~1, while aiming to maximize the full objective function in Stage~2. 

This strategy intentionally creates datasets of different sizes. The low-fidelity datasets (for example, used for $\mathbf{y}_\text{width}$) contain $n_{\text{total}}$ observations. In contrast, the high-fidelity datasets (for example, for $\mathbf{y}_{\text{front-strength}}$, $\mathbf{y}_{\text{back-strength}}$) contain $10 \cdot (n_{\text{total}}-n_{\text{Stage~1}})$ observations, where 10 is the number of strength measurement repetitions (see Table~\ref{tab:quality_params} for an overview of the qualities and constraints). By modeling the various physical properties with independent GPs, the framework efficiently handles these different dataset sizes. This approach is particularly advantageous for the high-fidelity objectives: by fitting the die strength GPs to all 10 raw data points from each configuration, the model's likelihood function, shown in Eq.~\eqref{eq:loglikelihood}, explicitly learns the high observation variance inherent in the destructive testing process.

This two-stage sequential approach can be interpreted as a pragmatic implementation of MF-BO~\citep{poloczek2017multi}. Our strategy leverages the low-cost, optical data to efficiently learn several objectives and constraints, and to identify promising sub-regions, while reserving the expensive, destructive tests for fine-tuning the primary quality objectives within that region.


\section{Experimental Methodology}\label{sec:5setup_experimental}
To validate the BOLD framework, we designed experiments on an industrial-grade platform, representative of a real-world semiconductor manufacturing environment.

\subsection{System and materials}
\begin{figure}[h] 
    \centering
    \newcommand{\myimgheight}{2.85cm}

    \begin{subfigure}[t]{0.24\textwidth} 
        \centering
        \includegraphics[height=\myimgheight]{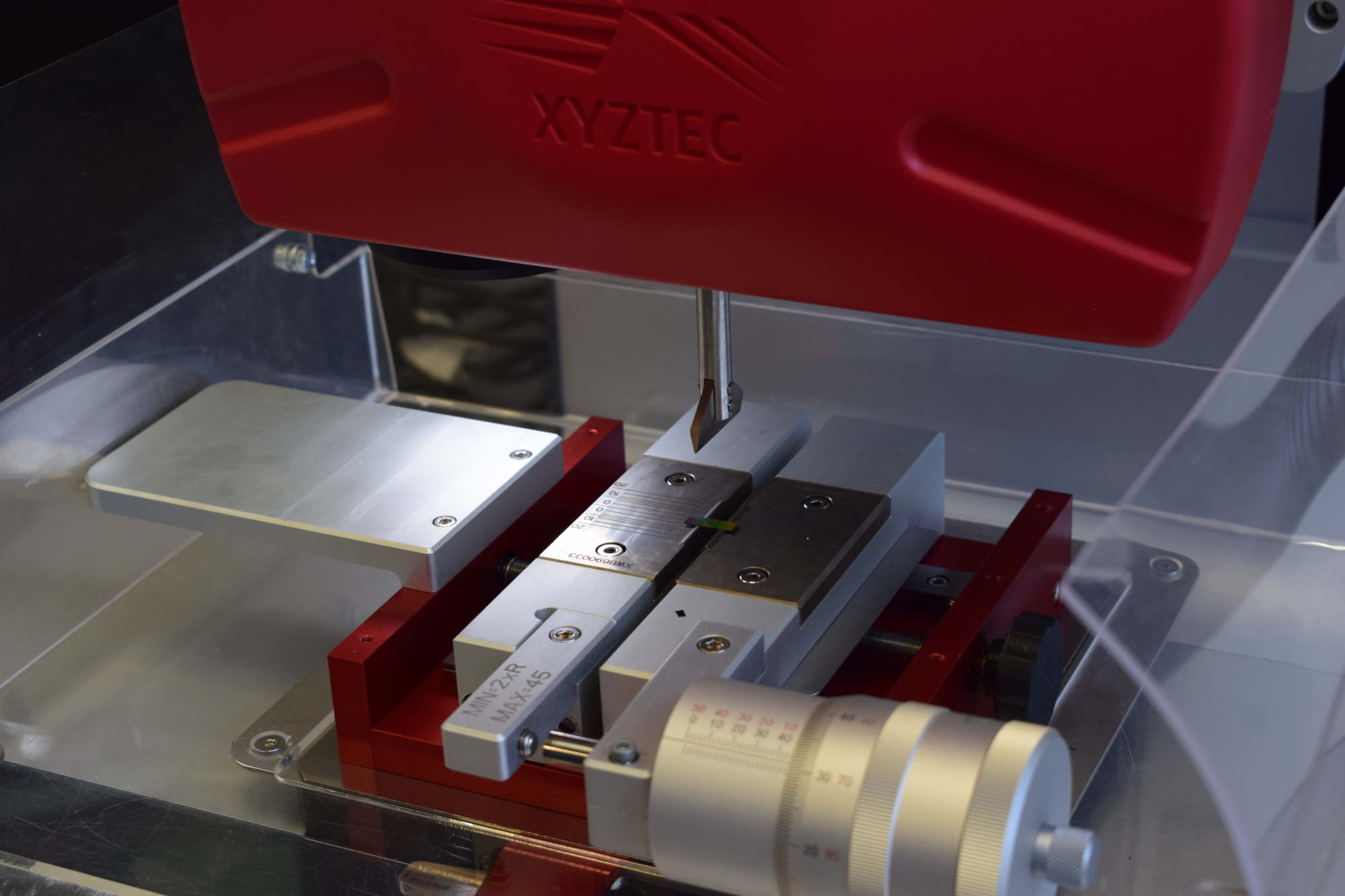}
        \label{fig:ds_cand1}
    \end{subfigure}%
    \hfill 
    \begin{subfigure}[t]{0.24\textwidth}
        \centering
        \includegraphics[height=\myimgheight]{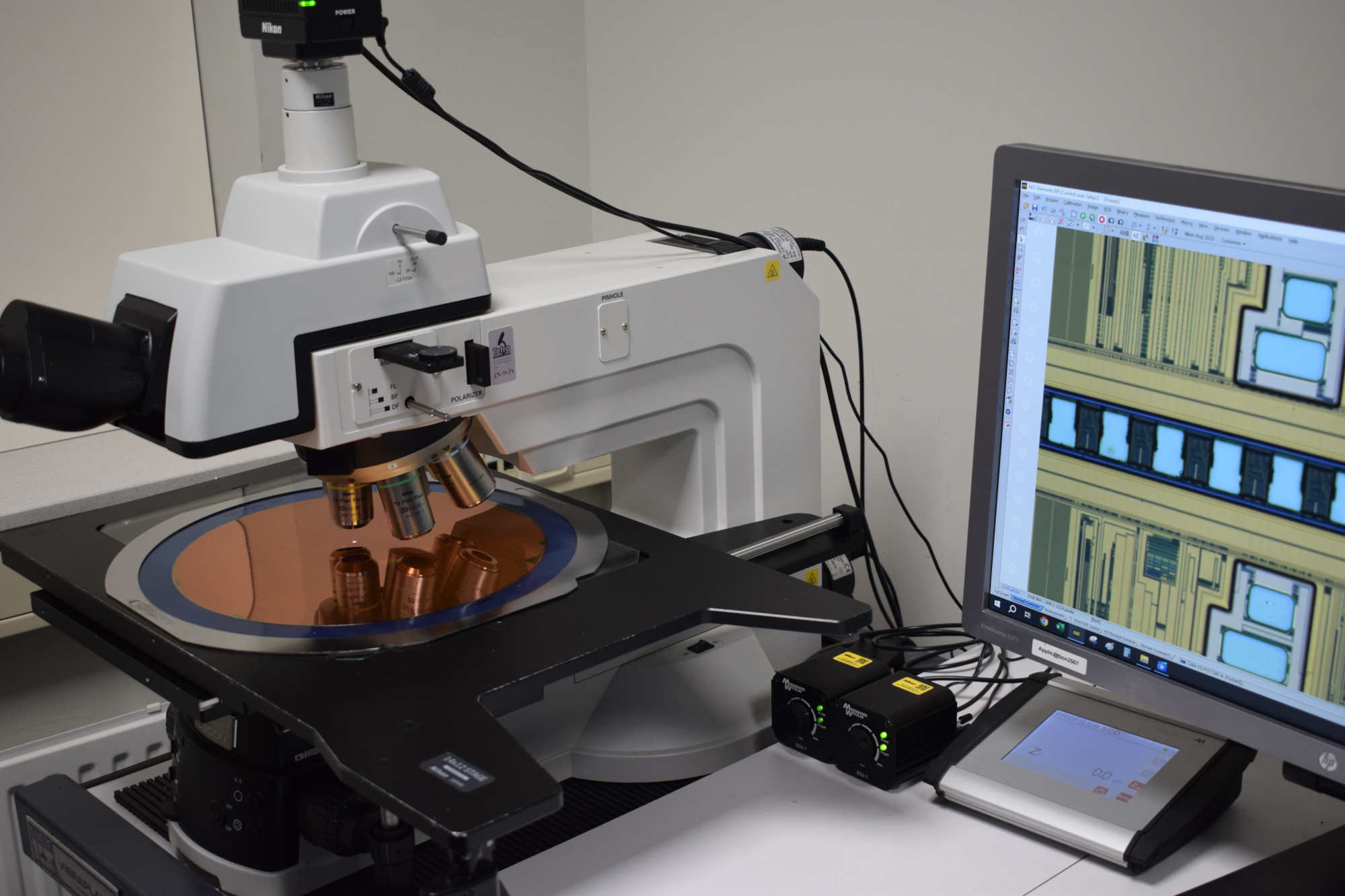}
        \label{fig:micro_cand1}
    \end{subfigure}%
    \hfill 
    \begin{subfigure}[t]{0.24\textwidth}
        \centering
        \includegraphics[height=\myimgheight]{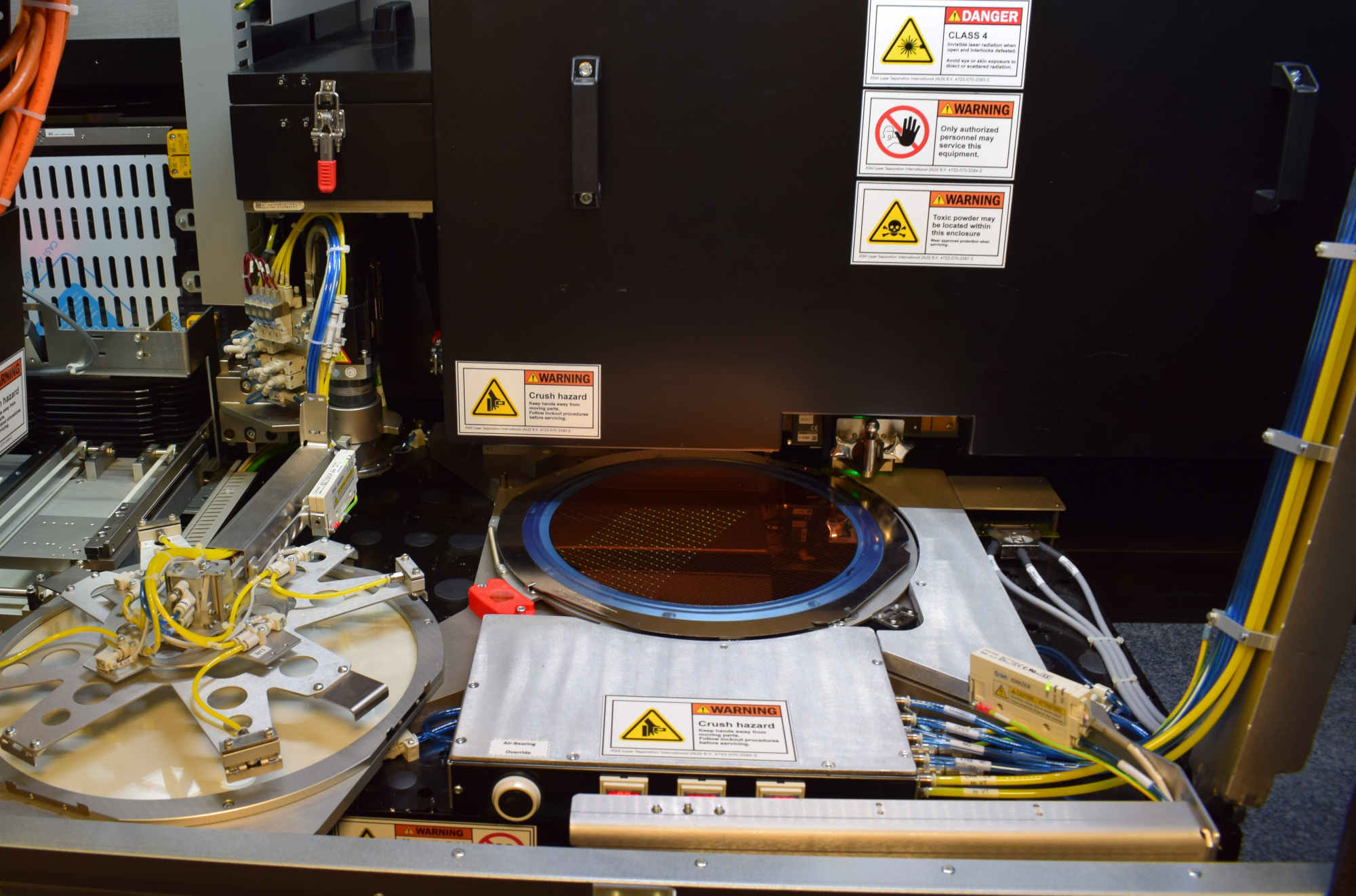}
        \label{fig:inside_cand82}
    \end{subfigure}%
    \hfill 
    \begin{subfigure}[t]{0.24\textwidth}
        \centering
        \includegraphics[height=\myimgheight]{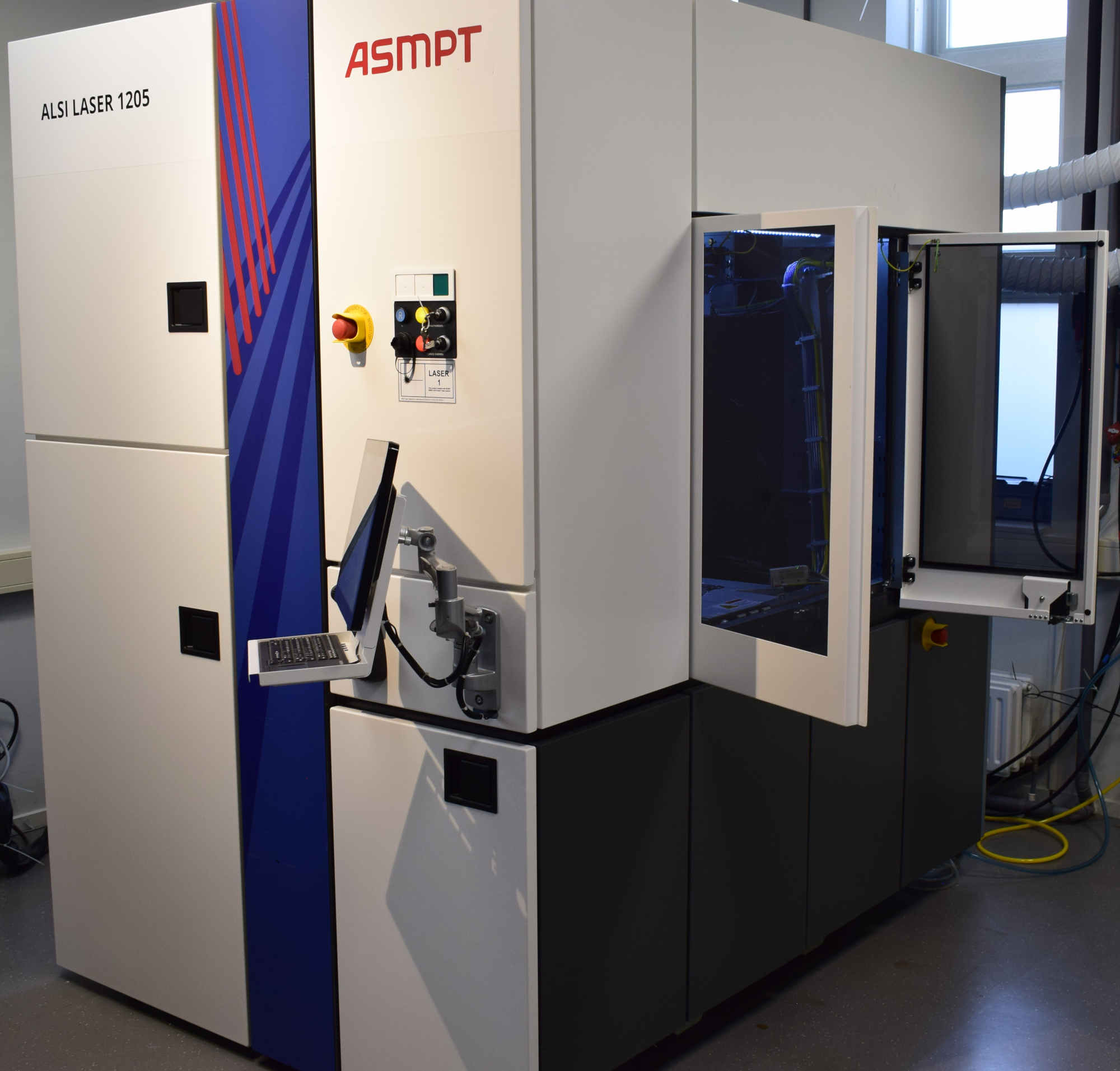}
        \label{fig:outside_cand74}
    \end{subfigure}
    \caption{\textbf{Experimental setup used throughout all experiments}. Die strength testing (\textbf{top left}), optical inspection (\textbf{top right}), together with the LASER1205 D-UVP system inside (\textbf{bottom left}) and outside (\textbf{bottom right}).}
    \label{fig:laser1205photos}
\end{figure}

Both experiments were conducted on an industrial LASER1205 D-UVP system, that leverages diffractive optical elements (DOEs) to split the primary laser beam into predefined patterns. The machine's capacity for up to eight distinct DOEs provides a large degree of control for processing different products and materials, essential for processing diverse and complex materials. Following each dicing process, non-destructive optical evaluation was performed using a Nikon Eclipse L300N microscope. Destructive mechanical validation was conducted using an XYZtec Condor Sigma three-point bending tester. The complete experimental apparatus is depicted in Fig.~\ref{fig:laser1205photos}.

We investigated two distinct wafer types: bare silicon wafers for the initial experiment, representing a baseline material, and production wafers in the second study. The bare silicon wafers are 60 \si{\micro\meter} thick, while in the second experiment 53 \si{\micro\meter} thick silicon product wafers were used. The top dielectric layers were around 8 \si{\micro\meter} thick, while the backside metal was around 4 \si{\micro\meter} thick. As is standard in industry, the precise material composition of the production wafers is proprietary and thus considered an unknown, black-box element of the optimization problem.

\subsection{Separation process specification}
We adopt a laser process design consisting of a three-pass laser dicing process, corresponding to a trenching, dicing and recovery pass. Each sequential step is executed by a different DOE: 
\begin{enumerate}
    \item \textbf{Trenching} employed a two-beam pattern to create the initial kerf.
    \item \textbf{Dicing} used a line of laser spots, distributing the energy along the laser cut.
    \item \textbf{Recovery} utilized a two-dimensional V-DOE pattern to treat the entire kerf edge, to ensure that the complete edge of the dicing kerf can be treated.
\end{enumerate}
These different beam patterns are designed and chosen to match each other, and reflect common design choices in laser dicing processes. 

The trenching pass was defined by three tunable parameters (power, step, angle), while the dicing and recovery passes were each defined by four (power, focus, frequency, step). Together, these 11 parameters allow for a wide range of process control, and contain several cross correlations. For example, the intensity (power/area) is influenced by the laser power and the focus position. As such, we optimize them simultaneously, forming the optimization space $\Omega \in \R^{11}$. Since the laser process can only be varied in between passes and not during a pass, this is a static black-box optimization problem. A complete list of all parameters, their bounds, and descriptions is provided in Table~\ref{tab:process_params}. 

\subsection{Measurement protocols}
To measure the required objective and constraint functions, we follow the following procedure. First, the optical measurements are measured by the technician, using software-aided visual inspection to estimate the burr height, the dicing width, and the modification width. Then, the percentage of chipouts and separation are measured by considering all the separated streets. Similarly, the cracks (front side, front side corner, backside, backside corner) are expressed in a percentage over all the processed material. 
For the destructive die strength measurements, ten tests are applied to the front side and ten tests to the back side strength. The mean (per category) over these samples is then used as a proxy to compute the objective score. For the validation study, we instead apply 15 strength tests on the front and on the back sides. A full overview and explanation of all measured objectives and constraint is provided in~\ref{app1}.

\section{Experimental Results}\label{sec:6results}
\subsection{Bare silicon wafer experiment}
\begin{table}[t] 
\centering
\begin{small}
\caption{
    \textbf{Bare Silicon Wafer Experiment}. Performance of the optimized process configuration (BOLD) compared to the engineering requirements. The method successfully met or exceeded all targets. Die strength is mean $\pm$ std. dev. over 10 measurements. Upward arrows indicates that higher values are desirable, whereas downward arrows indicate lower values are desirable.
}

\label{tab:experiment1-table-tall}
\begin{tabular}{l l l} 
\toprule
\textbf{Metric} & \textbf{Requirement} & \textbf{BOLD} \\
\midrule

\multicolumn{3}{l}{\textit{Performance Metrics}} \\ 
\quad Speed (\si{wafer/hr}) $\uparrow$ & $> 2.50 $ & 3.77 \\
\quad Strength (Front) (\si{\MPa}) $\uparrow$ & $ > 600$ & $671 \pm 57$ \\
\quad Strength (Back) (\si{\MPa}) $\uparrow$ & $ > 600$ & $613 \pm 40$ \\
\midrule

\multicolumn{3}{l}{\textit{Visual Quality}} \\
\quad Width (\si{\micro\meter}) $\downarrow$ & $[28.0 - 32.0]$ & 28.9 \\
\quad Burr height (\si{\micro\meter}) $\downarrow$ & $[0.0-2.5]$ & 2.0 \\
\midrule 

\multicolumn{3}{l}{\textit{Structural Integrity}} \\
\quad Cracks (Front) (\si{\percent}) $\downarrow$ & $[0-10]$ & 0 \\
\quad Cracks (Corner) (\si{\percent}) $\downarrow$ & $[0-10]$ & 0 \\
\quad Cracks (Back) (\si{\percent}) $\downarrow$ & $[0-10]$ & 0 \\
\quad Separation (\si{\percent}) $\uparrow$ & $100$ & 100 \\

\bottomrule
\end{tabular}
\end{small}
\end{table}

\begin{figure}[h]
    \centering
     \begin{subfigure}{0.475\linewidth}
         \centering
         \includegraphics[width=\textwidth]{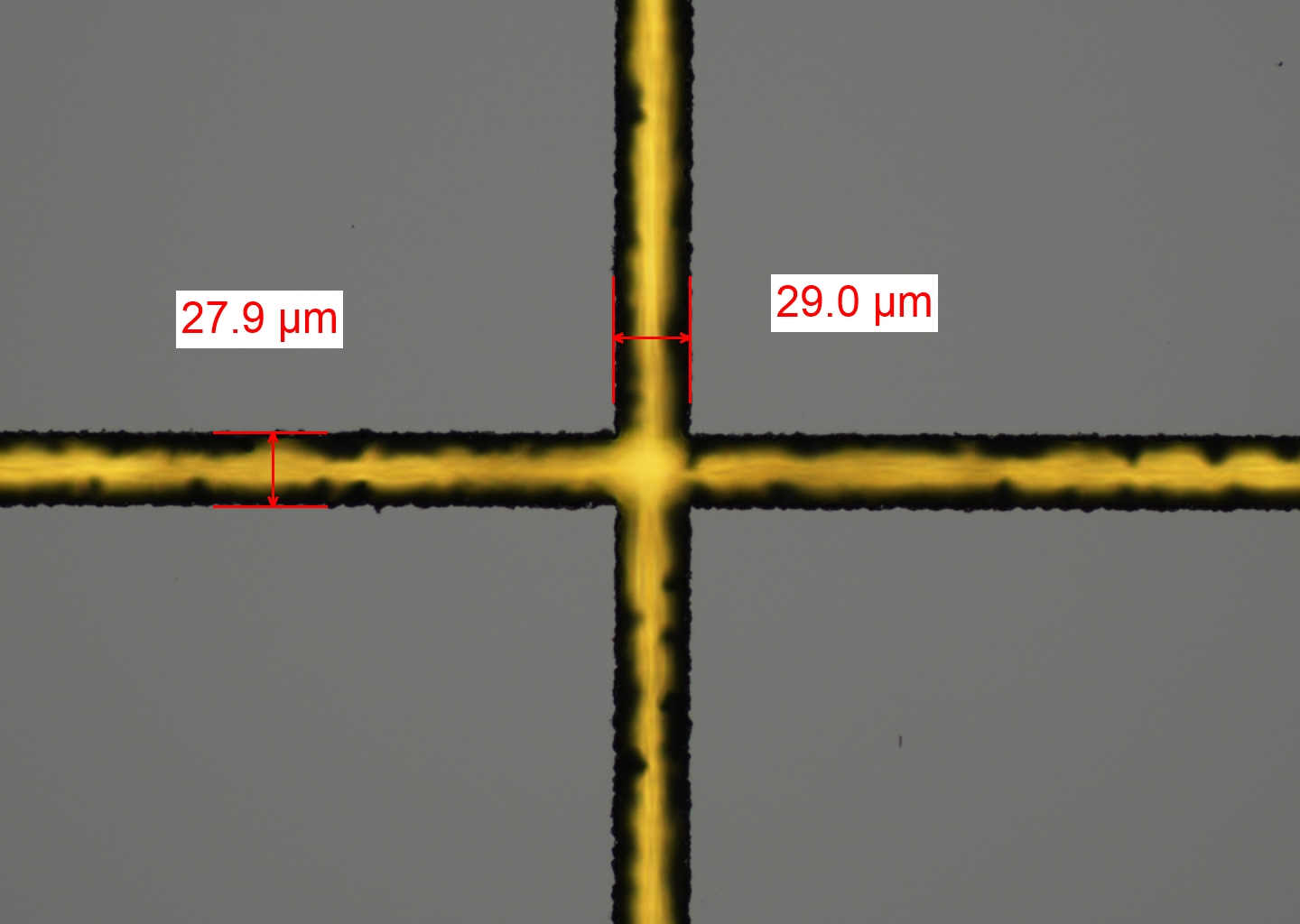}
        \label{fig:exp1-frontsidephoto}
    \end{subfigure}
      \begin{subfigure}{0.475\linewidth}
         \centering
         \includegraphics[width=\textwidth]{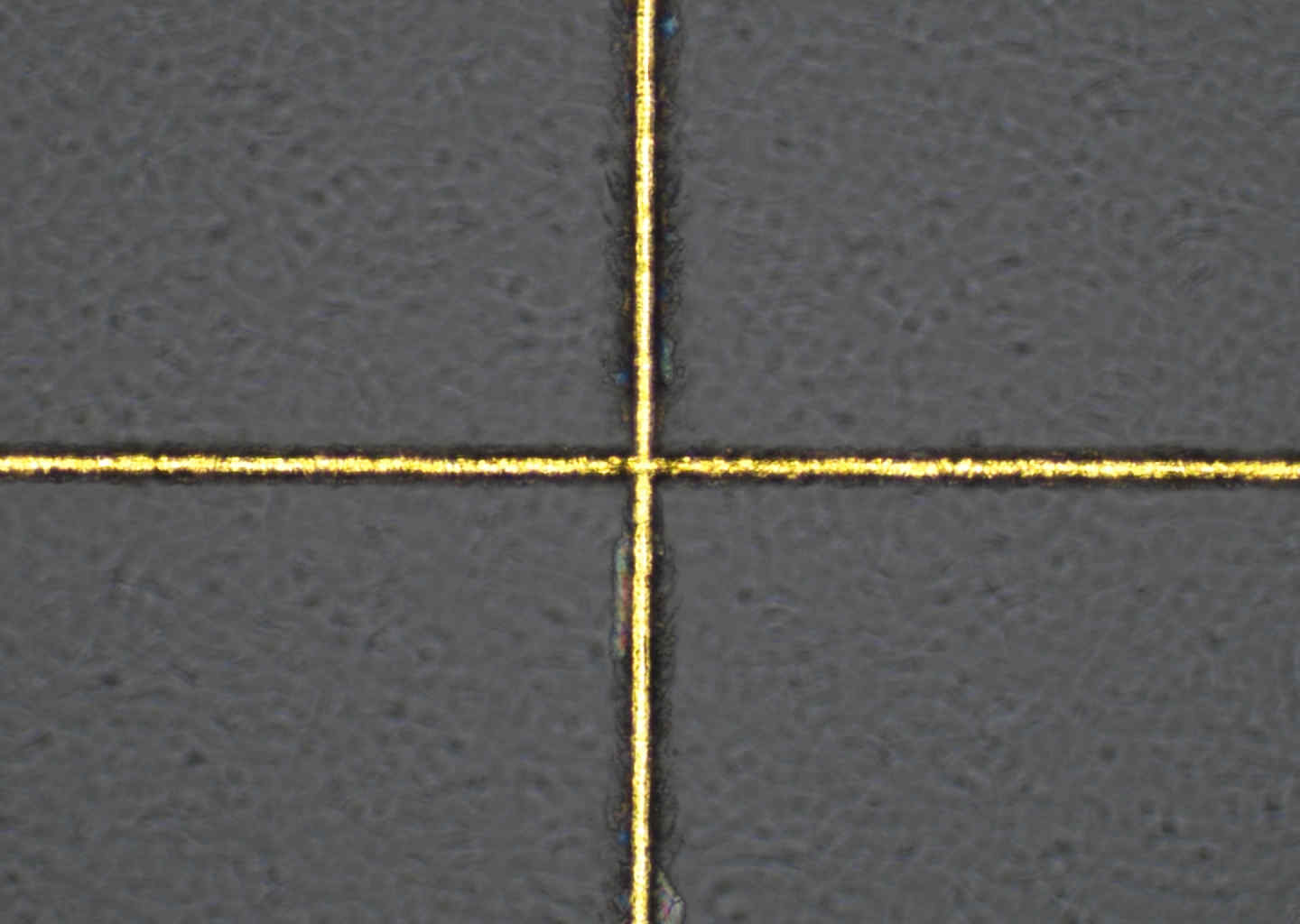}
         \label{fig:exp1-backsidephoto}
     \end{subfigure}
    \centering
    \caption{\textbf{Bare Silicon Wafer Experiment}. Frontside \textbf{(left)} and backside \textbf{(right) }of a processed die separated by the best configuration discovered by BOLD.}
    \label{fig:experiment1-photos}
\end{figure}

We first validated the BOLD framework on a 60 \si{\micro\meter} bare silicon wafer. This material serves as an essential baseline, representing a simpler case with lower material cost and complexity (that is, no lithographic top layers) compared to a full production wafer.

\subsubsection{Model configuration}
For this initial experiment, the expert-derived utility function was modeled directly with a single-output GP, rather than as a composite of independent GPs -- a simplification that was refined in the second experiment. The optimization used the constrained Thompson sampling framework with a batch size of $q=2$. The initial dataset consisted of 9 configurations: 4 standard database configurations and 5 generated via a Sobol sequence~\citep{SOBOL196786}. The learned constraint models $\mathbf{c}(\mathbf{x}) \sim \mathcal{GP}\left(\mu_0, k \mid \mathcal{D}_C\right)$ included front, corner, and back cracks, and separation, but not chipouts, which are not a relevant failure mode for this material. The expert-derived utility weights are described in Table~\ref{tab:utility-weights}. 

\begin{figure}[t]
    \centering
    \includegraphics[width=\linewidth]{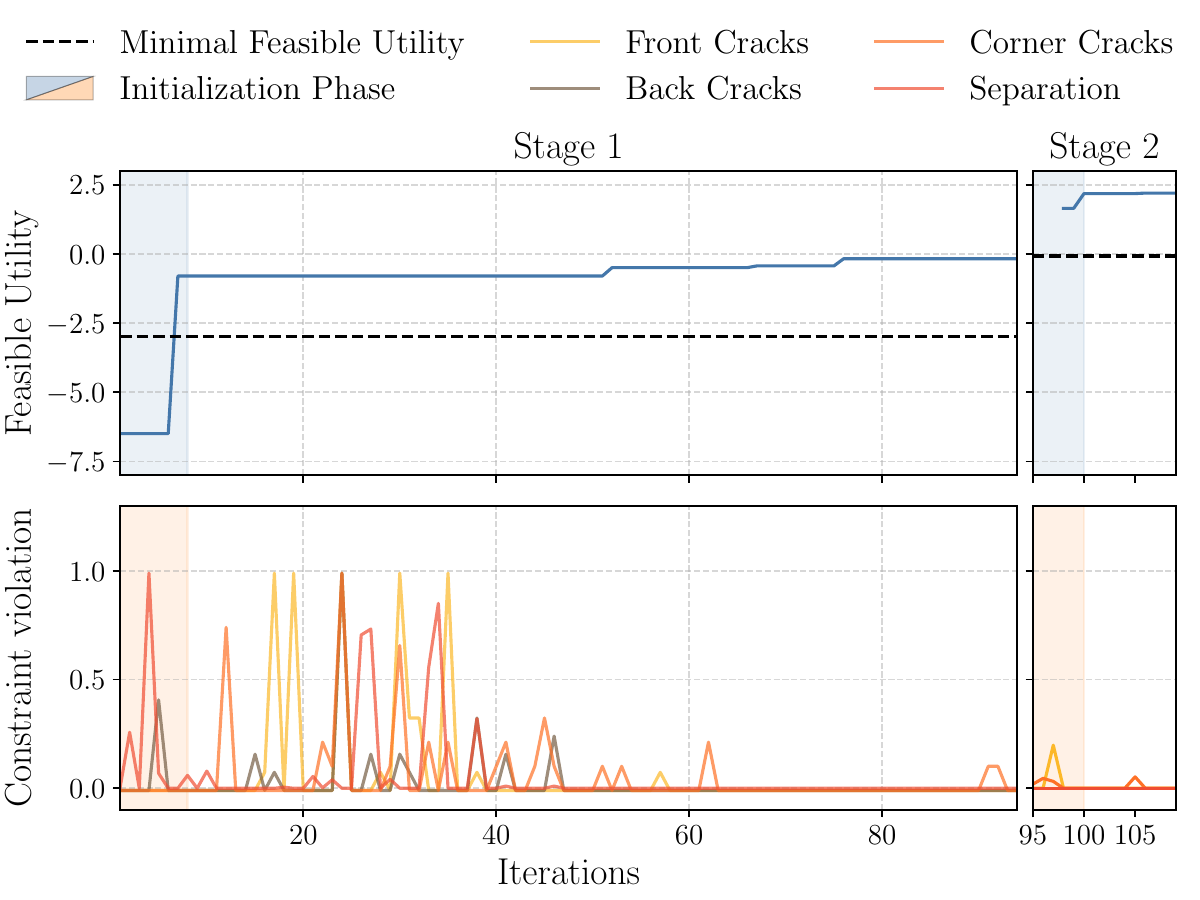}
    \caption{\textbf{Bare Silicon Experiment Convergence}. Performance of BOLD over the two stages (optical for Stage~1, and optical and destructive for Stage~2). \textbf{(Top)} Best-so-far feasible utility achieved during the experiment. The discontinuity at iteration 97 marks the switch to the Stage 2 objective function. \textbf{(Bottom)} Individual constraint violations (front, corner, back cracks, and separation) per iteration, demonstrating the algorithm's active exploration and subsequent learning to avoid infeasible solutions.}
    \label{fig:experiment1-convergenceplot}
\end{figure}
\begin{figure}[th]
    \centering
    \includegraphics[width=\linewidth]{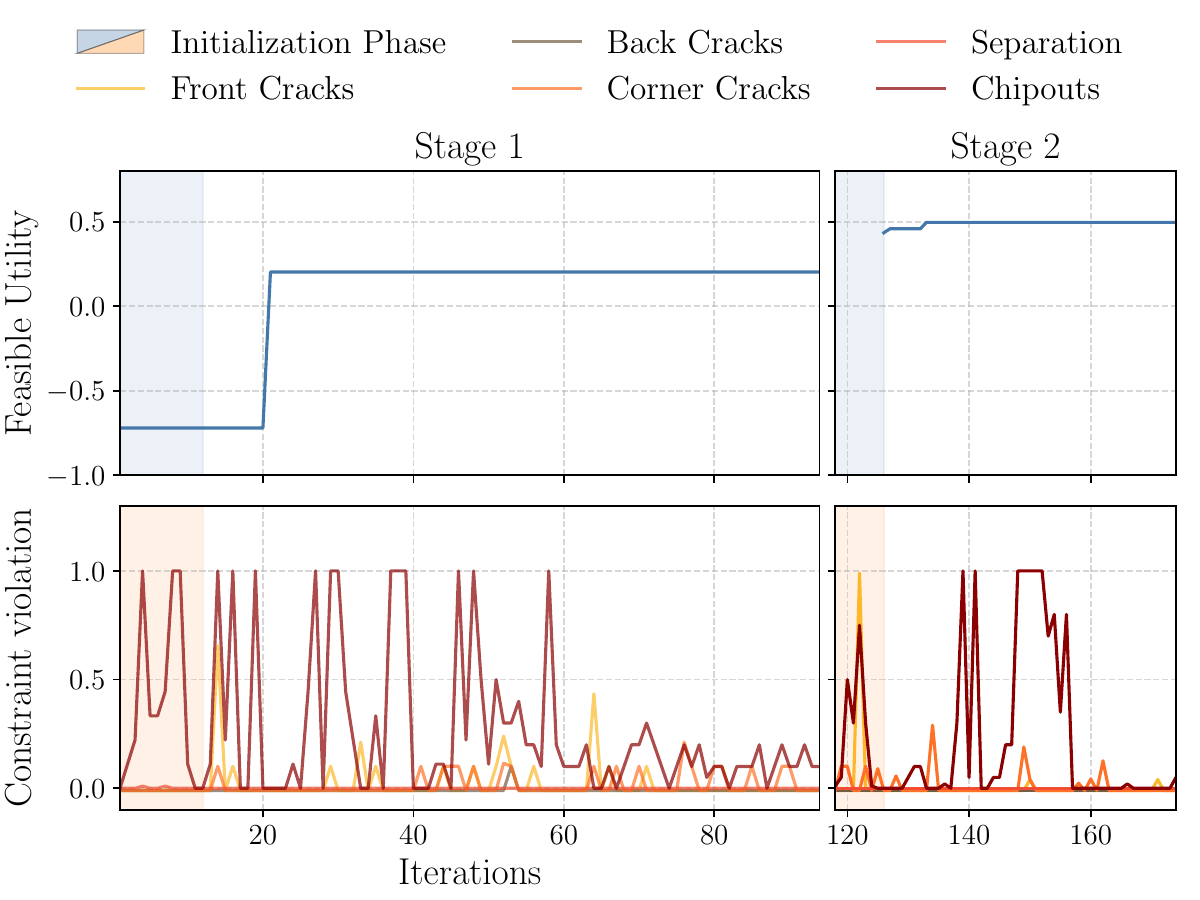}
    \caption{\textbf{Production Wafer Experiment Convergence}. Performance of BOLD over the two stages (optical for Stage~1, and optical and destructive for Stage~2). \textbf{(Top)} Best-so-far feasible utility achieved during the experiment. The discontinuity at iteration 117 marks the switch to the Stage~2 objective function. \textbf{(Bottom)} Individual constraint violations (front, corner, back cracks, separation, and chipouts) per iteration. The optimization is dominated by the chipouts constraint (dark red) -- a challenge specific to production wafers.}
    \label{fig:experiment2-convergenceplot}
\end{figure}
\subsubsection{Optimization performance}
The optimization progression is shown in Fig.~\ref{fig:experiment1-convergenceplot}, while Fig.~\ref{fig:experiment1-appendix} depicts the convergence of modification width, burr height, and productivity throughout the experiment. Following the two-stage procedure, the initial stage (optical measurements) ran for 94 iterations, while the second stage (optical and destructive measurements) ran for 15 iterations. The total experimental time was approximately three weeks. In total, two wafers were consumed for the initial stage and one wafer was consumed for the second stage. 

The bottom panel of Fig.~\ref{fig:experiment1-convergenceplot} depicts the algorithm's active learning of constraints. In the first $\sim$25 iterations, BOLD actively explored the parameter space, frequently violating constraints. After this initial exploration, the GP constraint models became sufficiently accurate, and the algorithm learned to avoid these infeasible regions, driving the constraint violation to zero for the remainder of the experiment.
The top panel shows the best-so-far feasible utility, which increases rapidly and stabilized after iteration 40. The discontinuity at iteration 97 marks the switch to the full objective in the second stage. The algorithm quickly discovers a configuration that satisfied this new, more complex setup.

The best-found feasible configuration from this experiment was validated against the pre-defined engineering requirements. The results, summarized in Table~\ref{tab:experiment1-table-tall}, show that the BOLD-discovered configurations successfully met or exceeded all engineering targets. It achieved a production speed of 3.77 wafer/hr, and surpassed the required 600 MPa for both front and back die strength. All structural integrity and visual quality metrics were also within their required bounds. Figure~\ref{fig:experiment1-photos} depicts the front- and back-side of the separated material under this process configuration, showing a clean separation at the dicing streets on both sides of the wafer.

\subsection{Production wafer experiment}

\begin{table*}[t] 
    \centering
    \begin{small}
    \caption{\textbf{Production wafer experiment}. Performance comparison of all final configurations from Experiment II. (mean die strength $\pm$ std. dev. over 15 measurements, except for the BOLD configuration which is measured 10 times. Upward arrows indicates that higher values are desirable, whereas downward arrows indicate lower values are desirable.}
    \setlength{\tabcolsep}{4pt}
    \label{tab:experiment2_results}
    
    \begin{tabular}{l S[table-format=1.2] c c S[table-format=2.1] c c c c c c c}
    \toprule
    & & \multicolumn{2}{c}{Die Strength} & \multicolumn{3}{c}{Visual Quality} & \multicolumn{4}{c}{Structural Integrity} &  \\
    \cmidrule(lr){3-4} \cmidrule(lr){5-7} \cmidrule(lr){8-11} 
    Method & {Speed} & {Front} & {Back} & {Width} & {Burr H.} & {Chipouts} & {Front} & {Corner} & {Back} & {Sep.} & Feasible\\
    & {(\si{wafer/hr})$\uparrow$ } & {(\si{\MPa})$\uparrow$} & {(\si{\MPa})$\uparrow$} & {(\si{\micro\meter})$\downarrow$} & {(\si{\micro\meter})$\downarrow$} & {(\si{\percent})$\downarrow$} & {(\si{\percent})$\downarrow$} & {(\si{\percent})$\downarrow$} & {(\si{\percent})$\downarrow$} & {(\si{\percent})$\uparrow$} &  \\
    \midrule
    
    Database & 3.70 & $\mathbf{496} \pm 86$ & $397 \pm 47$ & 33.2 & $\mathbf{0.5}$ & 50 & 33 & 0 & 0 & 100 & \xmark\\
    Database + Expert & 3.01 & $443\pm 58$ &  $419\pm 21$ & $\mathbf{28.1}$ & $\mathbf{0.5}$ & $1$ &  0 &  0 &  0 & 100 & \cmark \\
    \midrule 
    
    BOLD  & 2.81 & $\mathbf{496} \pm50$ & $394 \pm 26$ &  28.9 & $\mathbf{0.5}$ & 2 & 4 & 0 &  0& 100 & \cmark \\
    BOLD-A & 3.25 & $475 \pm 60$ & $\mathbf{424} \pm 40$ & 29.6 & 1.0 & 10 & 0 & 0 & 0 & 100 & \cmark\\
    BOLD-B & 2.78 & $435 \pm 45$ & $409 \pm 48$ & 28.6 & 1.0 & 5  & 8 & 0 &   0& 100 & \cmark \\
    BOLD-C & 3.34 & $488\pm 46$ & $396 \pm 44$ & 31.7 & $\mathbf{0.5}$ & 5 &  0 &  16 & 0 & 100 & \xmark \\
    {\textbf{BOLD + Expert}} & {\textbf{4.03}} & $444\pm46$ & $388\pm 59$ & 31.6 & $\mathbf{0.5}$& 5 &  0  & 0  & 0 & 100 & \cmark\\
    
    \bottomrule
    \end{tabular}
    \end{small}
\end{table*}

The main experiment was conducted on 53 \si{\micro\meter} production wafers, which represent a significant increase in task complexity. Production wafers consist of multiple proprietary layers, including dielectric stacks and metal surfaces (as seen in Fig.~\ref{fig:experiment2-photos}). A successful configuration must be robust to these material variations.

\subsubsection{Model configuration}
This experiment introduced three key methodological advancements over the bare silicon run. First, as discussed in Sec.~\ref{sec:4.5subsection}, the framework was upgraded to use independent Gaussian processes for each of the $M$ objectives. Second, the chipouts failure mode, a critical challenge for this material, was added as a new learned constraint. Finally, the GP models were augmented with informative priors on the kernel hyperparameters. These priors were derived from the posterior hyperparameters of the models trained in the bare silicon experiment, implementing a basic form of transfer learning to accelerate the optimization. The optimization was run with a batch size of $q=3$. The expert-derived utility weights are described in Table~\ref{tab:utility-weights}. 

\subsubsection{Validation analysis and baselines}
\begin{figure*}[t]
    \centering
     \begin{subfigure}{0.245\textwidth}
         \centering
         \includegraphics[width=\textwidth]{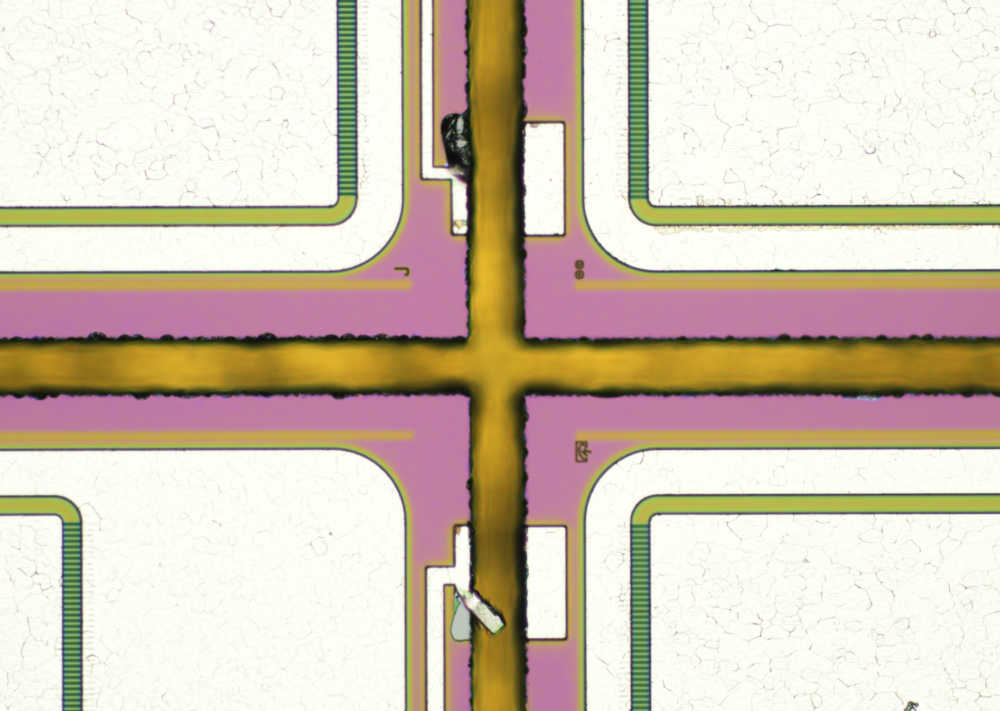}
         \caption{Database}
         \label{fig:validation-database}
     \end{subfigure}
     \hfill
     \begin{subfigure}{0.245\textwidth}
         \centering
         \includegraphics[width=\textwidth]{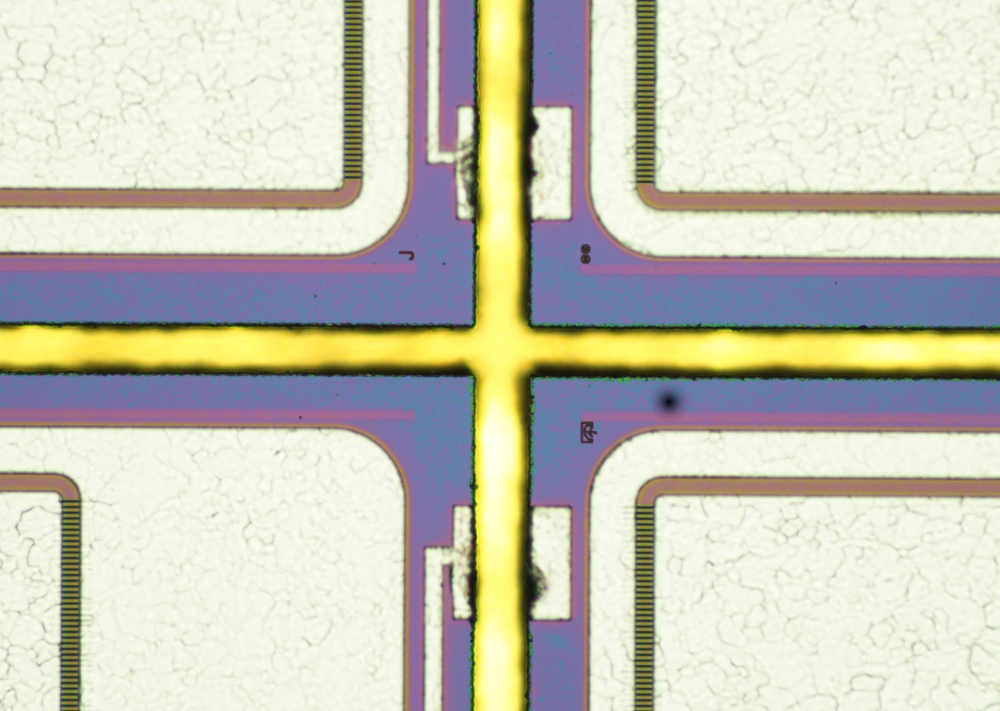}
         \caption{Database + Expert}
         \label{fig:validation-database+expert}
     \end{subfigure}
     \hfill
      \begin{subfigure}{0.245\textwidth}
         \centering
         \includegraphics[width=\textwidth]{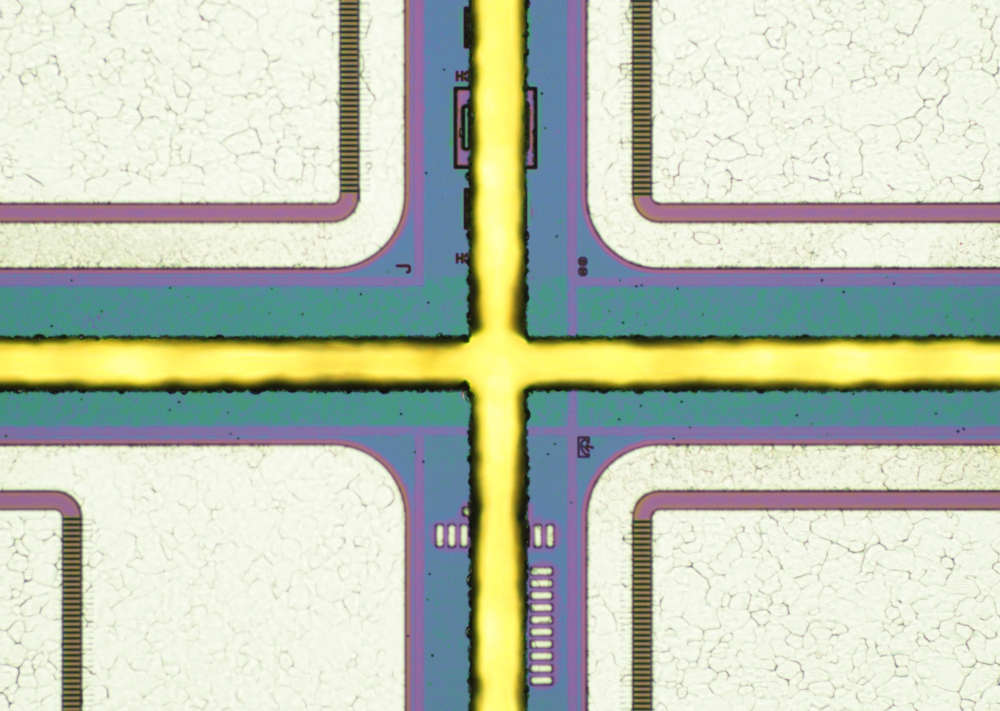}
         \caption{BOLD}
         \label{fig:validation-56}
     \end{subfigure}
     \hfill
      \begin{subfigure}{0.245\textwidth}
         \centering
         \includegraphics[width=\textwidth]{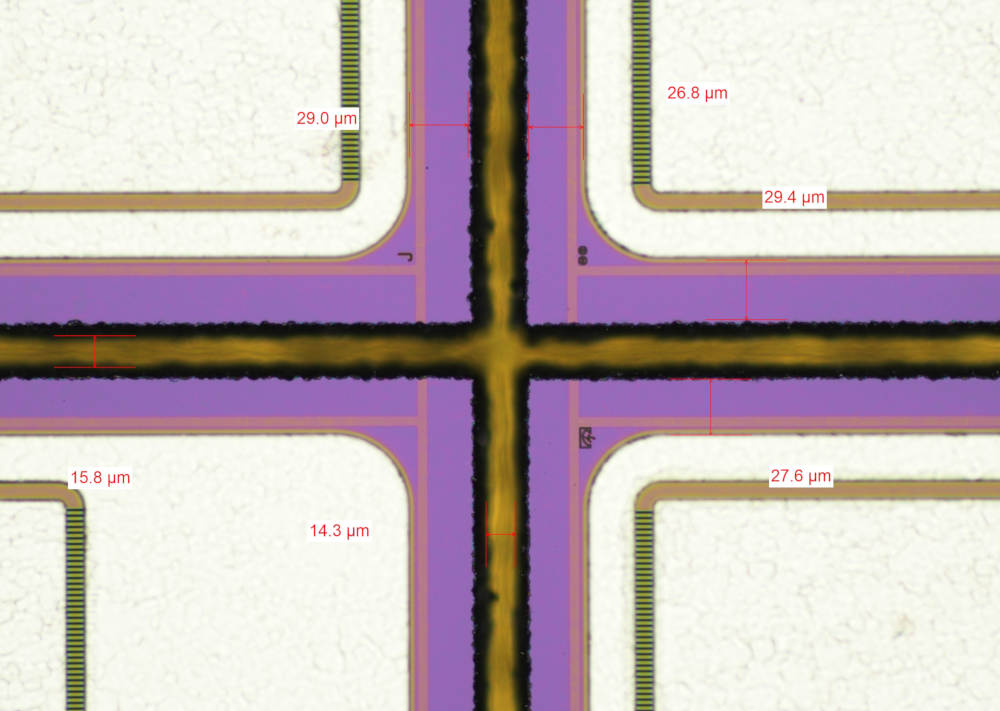}
         \caption{BOLD-A}
         \label{fig:validation-A}
     \end{subfigure}
     \begin{subfigure}{0.245\textwidth}
         \centering
         \includegraphics[width=\textwidth]{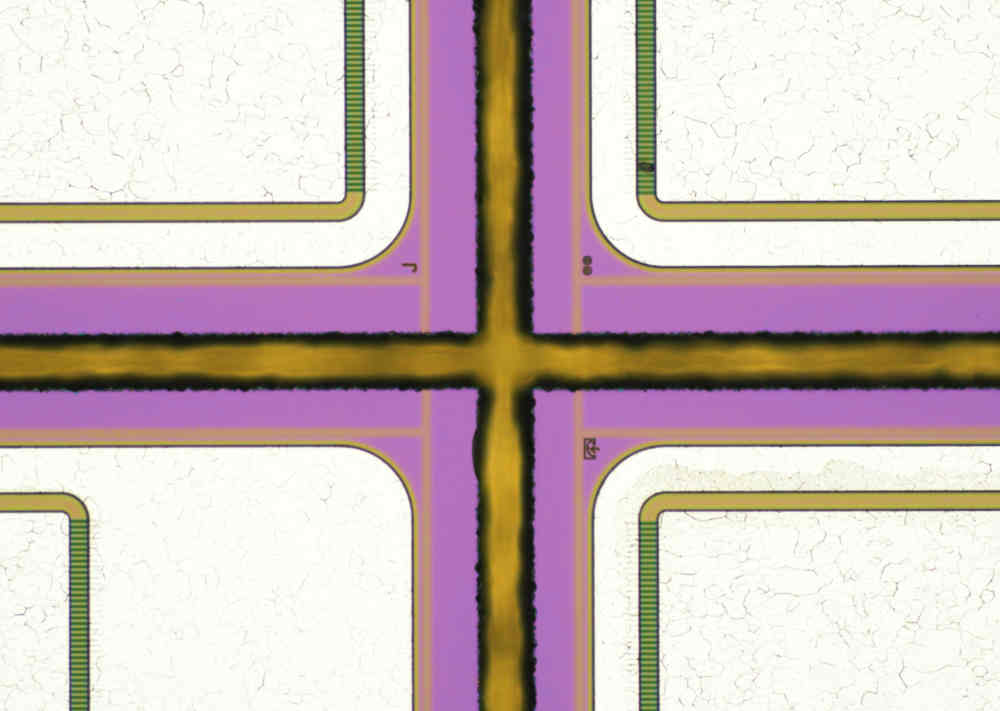}
         \caption{BOLD-B}
         \label{fig:validation-B}
     \end{subfigure}
     \kern0.25mm 
     \begin{subfigure}{0.245\textwidth}
         \centering
         \includegraphics[width=\textwidth]{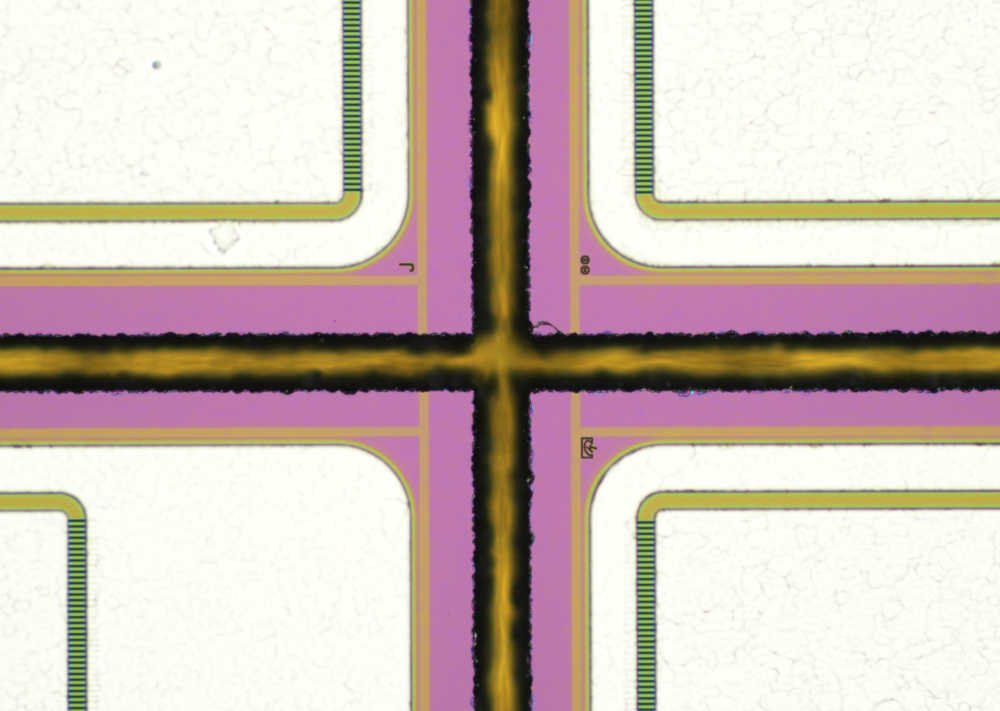}
         \caption{BOLD-C}
         \label{fig:validation-C}
     \end{subfigure}
     \kern0.25mm 
     \begin{subfigure}[b]{0.245\textwidth}
     \centering
     \includegraphics[width=\textwidth]{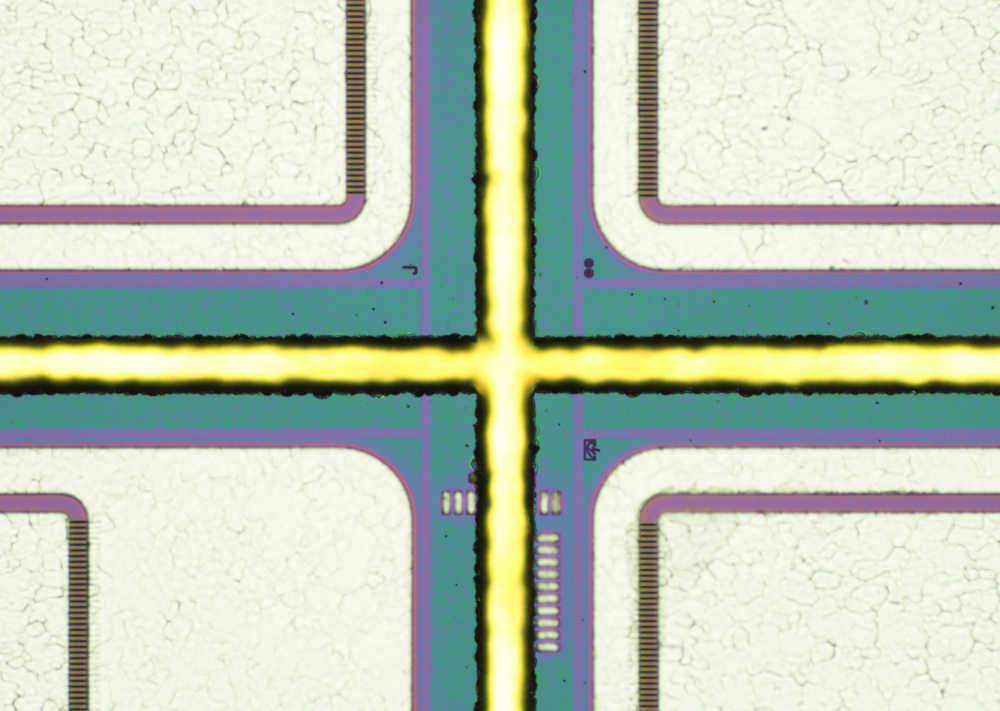}
     \caption{BOLD + Expert}
     \label{fig:validation-Expert+AI}
     \end{subfigure}
     \hfill
    \centering
    \caption{\textbf{Qualitative visual comparison of dicing results on the front-side of production wafers}. (a) The infeasible database configuration shows severe chipouts near metal structures. (b) The DB + Expert configuration provides a clean, high-quality baseline. (c-g) All BOLD-derived configurations successfully eliminate the chipouts. Notably, configuration (f) BOLD-C appears visually clean but is quantitatively infeasible due to violating structural integrity requirements (corner cracks, see Table~\ref{tab:experiment2_results}). The difference in colors across images is due to small differences in top layer thickness; variations that are common in product wafers.}
    \label{fig:experiment2-photos}
\end{figure*}

To validate the final configurations from the production wafer experiment, a comprehensive validation study was performed. Each configuration was executed on a single new, previously unused wafer to eliminate any potential variability in different wafer partials. To ensure high statistical confidence, the die strength validation was increased to 15 independent three-point bending tests for both the front and back sides (compared to 10 used during the optimization loop). The quantitative performance of the discovered configurations is presented in Table~\ref{tab:experiment2_results}, with top and bottom sides of the processed dies shown in Fig.~\ref{fig:experiment2-photos}. We evaluated three categories of configurations: industrial baselines, autonomous BOLD configurations, and lastly a synthesis between the BOLD configurations and human-in-the-loop approach.

The industrial baselines represent the standard, non-BO approach. The Database configuration is a pre-existing, off-the-shelf configuration known to be effective for other materials and products, and serving as the common starting point for new process development. This is compared to the `Database + Expert' baseline, where a process engineer took the `Database' configuration as a starting point and iteratively refined it by following established, rules-based guidelines according to standard industrial practice.

The autonomous BOLD configurations include several variants. The BOLD configuration is the best-found feasible configuration discovered during the optimization run. In contrast, BOLD-A is the maximum a posteriori (MAP) estimate, inferred from the final posterior distribution using the full dataset and the original expert-derived utility weights. Because the framework models all physical processes with independent posteriors, we also investigated the model's ability to generalize to different utility functions. This is a non-trivial task, as the resulting optima may lie in different parts of the input space than the data seen thus far, effectively testing out-of-distribution generalization. We generated two such configurations post-hoc: BOLD-B, which used weights strongly prioritizing die strength, and BOLD-C, which strongly prioritized production speed. Both were obtained by finding the MAP solution for their respective utility weights without any new experiments, and can be interpreted as a basic approximation of the Pareto front. We overview all utility function weights in Table~\ref{tab:utility-weights}.

Lastly, we investigate a human-in-the-loop approach `BOLD + Expert', where a process engineer used the BOLD-derived BOLD-A configuration as a new, high-quality starting point for one round of manual refinement, rather than starting from the Database.

\subsubsection{Optimization performance}
The optimization progression is detailed in Fig.~\ref{fig:experiment2-convergenceplot}, while Fig.~\ref{fig:experiment2-appendix} depicts the convergence of modification width, burr height, and productivity throughout the experiment. The experiment consisted of 117 iterations in the initial stage, consuming two wafers, and 57 iterations in the second stage, consuming another two wafers. The total experimental time was approximately three weeks, comparable to the time allotted for a manual process engineer (2-4 weeks, with 2-3 wafers). 

The convergence plots reveal a much more challenging optimization problem than the bare silicon experiment. The bottom panel of Fig.~\ref{fig:experiment2-convergenceplot} shows that the optimization was almost entirely dominated by a single constraint: chipouts (dark red). The algorithm persistently explored regions that caused this failure mode, demonstrating the difficulty of finding a feasible region in this 11-dimensional space. The BOLD framework successfully learned the boundary of this constraint without die strength measures in the initial stage after 117 iterations. After the initial measurements in Stage~2, the algorithm explores various configurations that can potentially improve the die strengths, leading to multiple constraint violations before slightly improving the best-so-far utility.

The top panel shows the best-so-far feasible utility. After a feasible solution is found in the initial stage, the best-so-far utility remains constant for many iterations due to the constraint violations in these iterations (making any utility scores ineligible). After the switch to the second stage, the best-so-far feasible utility increases again due to the high-fidelity objective at iteration 117. 

\subsubsection{Validation results}
As seen in Table~\ref{tab:experiment2_results}, the baseline achieves a high production speed, but this comes at the price of not meeting the product requirements: the processed wafers suffer from severe chipouts (50\% of the streets), and contain front material cracks in 33\% of the dies. The Database + Expert baseline forms a strong, feasible solution. The expert successfully eliminated all structural defects (Fig.~\ref{fig:validation-database+expert}), but this required a reduction in production speed to 3.01 wafer/hr. This feasible configuration, with strong die strength (443 MPa front, 419 MPa back), serves as the primary benchmark.

The (autonomous) BOLD configurations demonstrate the framework's success in discovering high-quality and feasible process configurations. The BOLD, BOLD-A, and BOLD-B configurations all successfully eliminated the critical chipout failures (Fig.~\ref{fig:validation-56},~\ref{fig:validation-B} and~\ref{fig:validation-C}) and were validated as feasible. Notably, the BOLD-A process not only met all product requirements but also achieved a production speed of 3.25 wafer/hr, an 8\% improvement over the expert baseline while maintaining comparable die strength. The post-hoc analysis shows that while the speed-prioritizing (BOLD-C) configuration was correctly identified by the model as fast (3.34 wafer/hr), it turned out to be infeasible, due to a 16\% corner crack rate (Fig.~\ref{fig:validation-C}). This demonstrates the critical role of the learned constraint models in identifying failure modes.

Finally, the BOLD + Expert configuration demonstrates the power of a human-in-the-loop approach. By using the BOLD-discovered BOLD-A as a starting point for one final round of expert refinement, the production speed was pushed to 4.03 wafer/hr. This represents a 34\% improvement over the original expert-only baseline, while still maintaining all quality and die strength requirements. The BOLD framework successfully navigated the challenging chipout constraint, finding a new, high-performance region of the parameter space that the expert could then further improve.

\subsection{Statistical analysis and model evaluations}
\begin{figure*}[t]
    \centering
    \includegraphics[width=\linewidth]{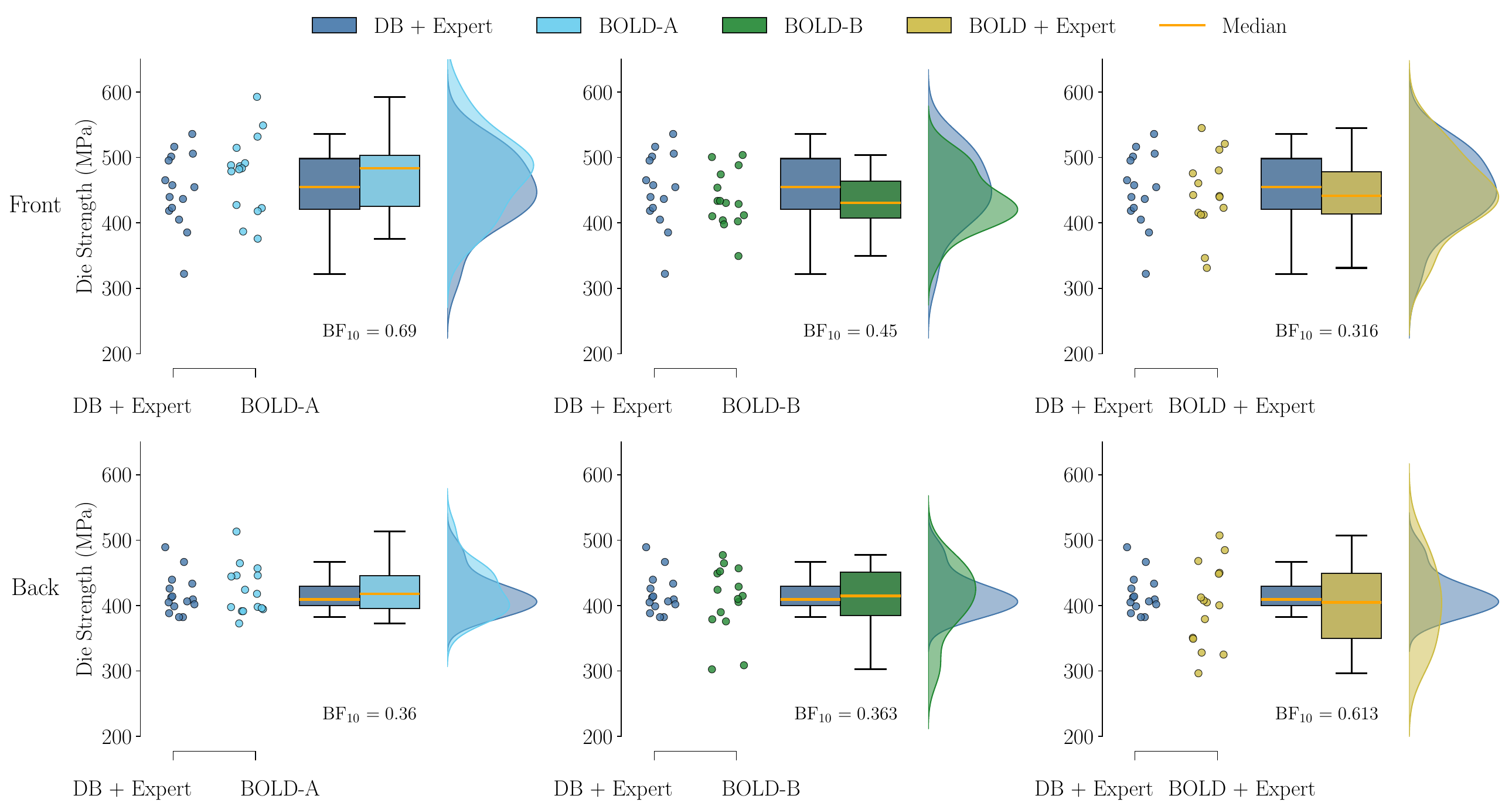}
    \caption{\textbf{Production wafer experiment.} Validation set die strength performance distributions for the feasible expert baseline and the BOLD-derived configurations. 
    In each category, the left box-plot shows the front die strength (solid) and its right counterpart the back die strength (dotted). The Bayes factors corresponding to the analysis in Table~\ref{tab:bayesianPairedSamplesT-Test} is annotated above the x-axis.}
    \label{fig:diestrength-boxplot-with-speed}
\end{figure*}
To statistically validate the final results, we compared the die strength of the Database + Expert configuration (our feasible baseline) against the three new feasible process configurations: BOLD-A, BOLD-B, and BOLD + Expert. We applied a Bayesian independent samples student's t-test. This method performs a Bayesian model comparison between two competing hypotheses: H$_0$, which states that there is no difference in the mean die strength between the two processes, and the alternative hypothesis H$_1$, which claims a difference in the means exists~\citep{jeffreys1998theory}. The Bayes Factor, BF$_{10}$, quantifies the evidence for H$_1$ relative to H$_0$. A key property of this approach is that it can quantify evidence for the null hypothesis (when BF$_{10} < 1$). Values below 1/3 are considered moderate evidence for H$_0$~\citep{lee2014bayesian}. We performed this comparison for both front and back die strength, for a total of six statistical tests.

The results are presented in Table~\ref{tab:bayesianPairedSamplesT-Test} and visualized in Fig.~\ref{fig:diestrength-boxplot-with-speed}. For all six comparisons, the Bayes Factor BF$_{10}$ was less than 1, indicating that the data supports the null hypothesis (no difference) over the alternative (a difference) for the given number of measurements. For front die strength, the BOLD + Expert configuration (BF$_{10}=0.316$) provided moderate evidence for the null hypothesis, suggesting no meaningful difference in mean strength compared to the expert baseline. Similarly, BOLD-B (BF$_{10}=0.450$) and BOLD-A (BF$_{10}=0.690$) also favored H$_0$. For back die strength, the evidence for equivalence was also clear. BOLD-A (BF$_{10}=0.360$) and BOLD-B (BF$_{10}=0.363$) both approached the moderate evidence threshold. These statistical findings are visually confirmed in Fig.~\ref{fig:diestrength-boxplot-with-speed}, where the box plots and raincloud distributions for the BOLD-derived configurations show significant overlap with the expert baseline, with their medians and mean remaining very close.

This analysis provides further evidence that the BOLD-derived process configurations, particularly the BOLD + Expert configuration, achieved significant speed improvements (up to 34\%) without sacrificing the critical engineering requirement of die strength.

\begin{table}[h]
    \centering
    \begin{small}
    \caption{\textbf{Bayesian Independent Samples T-Test}. Expert Baseline (Database + Expert) vs. feasible BOLD process configurations (BOLD-A, BOLD-B and BOLD + Expert). The Bayes Factor (BF$_{10}$) quantifies the evidence for the alternative hypothesis (H$_1$: the means are different). Bayes factors $<1$ support the null hypothesis (H$_0$: no difference), with values $<1/3$ considered moderate evidence in favor of the null hypothesis.}
    \label{tab:bayesianPairedSamplesT-Test}
    \begin{tabular}{l c c}
        \toprule
        & \multicolumn{2}{c}{BF$_{10}$ (vs. Expert Baseline)} \\
        \cmidrule(lr){2-3}
        Configuration & Front Die Strength & Back Die Strength \\
        \midrule
        BOLD-A        & 0.690 & 0.360 \\
        BOLD-B        & 0.450 & 0.363 \\
        BOLD + Expert & 0.316 & 0.613 \\
        \bottomrule
    \end{tabular}
    \end{small}
\end{table}


\section{Discussion}
\label{sec:7discussion}

\subsection{Autonomous discovery in complex industrial environments}
The primary contribution of this work is the demonstration that Bayesian optimization can autonomously navigate the complex, high-dimensional landscape of laser dicing processes. On complex product wafers, the BOLD framework successfully identified feasible process configurations that met all strict engineering requirements, effectively solving a task that typically takes weeks of expert effort, while requiring only technician-level oversight for comparable time and material consumption.

Crucially, the autonomous BOLD-A configuration not only achieved feasibility but also outperformed the manual expert baseline in production speed (3.25 vs 3.01 wafer/hr) while maintaining comparable die strength. This validates the core premise of our framework: that a data-driven approach can handle the non-linear interactions of 11 process parameters to find production-ready solutions without human intervention. 
Furthermore, the analysis of the discovered parameters revealed that the model identified effective V-DOE step sizes that contradicted common engineering heuristics. This indicates that the automated framework is capable of discovering novel, high-performance configurations that may be overlooked by manual tuning. 

While the framework is designed for autonomous discovery, our results indicate it also serves as a powerful tool for expert augmentation. The BOLD + Expert configuration demonstrated that a process engineer could leverage the AI-discovered solution as a high-quality starting point to push performance even further (reaching 4.03 wafer/hr). This suggests that BOLD can be deployed either as a fully autonomous solver or as a tool to accelerate expert workflows. The latter aligns closely with recent developments in human-in-the-loop machine learning, a paradigm designed to complement algorithmic efficiency with human expertise to enhance overall system performance~\citep{mosqueira2023human}.

\subsection{Managing constraints and evaluation costs}
The success of the autonomous optimization relied on addressing the specific constraints of the industrial environment. The progression of the product wafer experiment (Fig.~\ref{fig:experiment2-convergenceplot}) highlights that process discovery is fundamentally a constrained optimization problem; the algorithm spent the majority of the exploration phase learning to avoid chipout failure modes. The use of Gaussian process constraint models is critical here for efficiently identifying feasible operating regions within the search space.
Additionally, the sequential two-stage fidelity strategy proved essential for the method's practical viability. By utilizing fast optical measurements to filter the search space before introducing expensive destructive testing, the framework is able to drastically reduce material consumption.

\subsection{Future work and broader impact}
Future work could improve the autonomy of the system further. First, the current framework relies on a pre-determined Diffractive Optical Element (DOE). Future work could incorporate the choice of DOE directly into the optimization loop, creating a \textit{mixed-integer optimization} problem~\citep{daxberger2019mixed} that co-designs the optical hardware configuration alongside the process parameters. 
Second, the current utility function relies on fixed expert weights. Extending this to \textit{multi-objective optimization} would allow the algorithm to autonomously present a set of Pareto-optimal trade-offs between speed and quality~\citep{daulton2020differentiable}. Third, to further reduce the development time for new products, \textit{transfer learning} could be implemented to learn shared representations between data of different yet related wafer materials~\citep{bai2023transfer}. This would increase data efficiency, potentially bringing the optimization timeline down from weeks to days.

Beyond laser dicing, the principles of this framework are applicable to a wide range of complex industrial automation problems. By formalizing the problem as a constrained, high-dimensional black-box optimization task, BOLD provides a blueprint for converting heuristic-based manufacturing steps into fully autonomous, data-driven processes. This blueprint transfers naturally to related high-tech sectors such as robotics and renewable energy, where safe, efficient exploration of complex physical dynamics is critical. Beyond industrial automation, these principles are equally relevant to critical domains such as personalized healthcare, where the tuning of complex treatments can be turned into a precise, optimization-based science.

\section{Conclusion}
\label{sec:8conclusion}
In this work, we introduced BOLD, a framework for the automated discovery of laser dicing processes. By integrating trust-region Bayesian optimization, non-linear constraint learning, and a two-stage fidelity evaluation strategy, we addressed the challenges of high-dimensionality and conflicting objectives inherent to semiconductor manufacturing.

Our experimental validation on industrial-scale equipment confirms that the framework can autonomously discover feasible, high-performance process configurations for complex multi-layered wafers. The method successfully matched or exceeded expert baselines in both production speed and material integrity, while strictly adhering to known, proprietary process constraints and unknown, and learned constraints caused by laser-material interactions. These results represent a significant step toward data-driven, autonomous process discovery, providing a viable path to reduce development time and unlock new, high-performance operating regimes in critical manufacturing applications.


\section*{CRediT authorship contribution statement}
\textbf{David Leeftink:} Conceptualization, Methodology, Software, Validation, Formal analysis, Investigation, Data Curation, Visualization, Project administration, Writing - Original Draft, Writing - Review \& Editing. 
\textbf{Roman Doll:} Investigation, Resources, Writing - Original Draft, Project administration
\textbf{Heleen Visserman:} Investigation
\textbf{Marco Post:} Formal analysis
\textbf{Faysal Boughorbel:} Funding acquisition
\textbf{Max Hinne:} Supervision, Conceptualization, Methodology, Writing - Review \& Editing
\textbf{Marcel van Gerven:} Supervision, Conceptualization, Funding acquisition, Writing - Review \& Editing. 

\section*{Declaration of competing interest}
R. Doll, H. Visserman, M. Post, and F. Boughorbel are employed by ASM Pacific Technology. 
The authors declare that they have no other known competing financial interests or personal relationships that could have appeared to influence the work reported in this paper.

\section*{Acknowledgements}\label{sec:acknowledgements}
We thank Denis Deriga, Dr. Kees Biesheuvel, and Mark Mueller for their insightful comments on the laser process configuration design, the utility weights, and project execution. 

\section*{Funding}
This publication is part of the project ROBUST: Trustworthy AI-based Systems for Sustainable Growth with project number KICH3.LTP.20.006, which is (partly) financed by the Dutch Research Council (NWO), ASMPT, and the Dutch Ministry of Economic Affairs and Climate Policy (EZK) under the program LTP KIC 2020-2023. All content represents the opinion
of the authors, which is not necessarily shared or endorsed by their respective employers and/or sponsors.

\section*{Data availability}
The data that support the findings of this study are not publicly available due to their proprietary nature and commercial sensitivity regarding the industrial manufacturing process.

\appendix
\section{Process details} \label{app1}
The key laser process parameters and operating are depicted in Table~\ref{tab:process_params}, the quality parameters for the dicing process are shown in Table~\ref{tab:quality_params}, while the utility-function weight settings are described in Table~\ref{tab:utility-weights}.

\begin{table*}
    \centering
    \begin{small}
    \caption{Key laser process parameters and their operating constraints for each pass.}
    \label{tab:process_params}
    \begin{tabular}{lllll}
        \toprule
        \textbf{Laser Pass} & \textbf{Parameter} & \textbf{Step Size} & \textbf{Explanation} & \textbf{Operational Constraint} \\
        \midrule
        Trenching & Power [W] & 0.1 & Used power during trench & Max. guaranteed power \\
        & Step [\si{\micro\meter}] & 0.1 & Distance between two pulses & Inside known step range \\
        & RDU angle [Degree] & 0.2 & Rotation of DOE & Depending on used DOEs \\
        \midrule
        Dicing & Power [W] & 0.2 & Used power during cut & Max. guaranteed power \\
        & Focus [\si{\micro\meter}] & 10 & Focus position of laser spot & $\pm$ one Rayleigh length \\
        & Step [\si{\micro\meter}] & 0.1 & Distance between two pulses & Inside known step range \\
        & Frequency [Hz] & 1000 & \# laser pulses (during cut) & $<$ fixed limit \\
        \midrule
        Recovery & Power [W] & 0.2 & Used power during V-DOE & Max. guaranteed power \\
        & Focus [\si{\micro\meter}] & 10 & Focus position of laser spot & $\pm$ one Rayleigh length \\
        & Step [\si{\micro\meter}] & 0.1 & Distance between two pulses & Inside known step range \\
        & Frequency [Hz] & 1000 & \# laser pulses during V-DOE & $<$ fixed limit \\
        \bottomrule
    \end{tabular}
    \end{small}
\end{table*}

\begin{table*}
    \centering
    \begin{small}
    \caption{Objectives and constraints for the laser dicing process.}
    \label{tab:quality_params}
    \begin{tabular}{lllll}
        \toprule
        \textbf{Quality Parameter} & \textbf{Unit} & \textbf{Purpose} & \textbf{Explanation} & \textbf{Measurement Method}  \\
        \midrule
        Front die strength & [MPa] & Objective& Top side resistance against mechanical breaking & Destructive \\
        Front die strength & [MPa] & Objective& Back side resistance against mechanical breaking & Destructive \\
        Kerf width & [\SI{}{\micro\meter}]  & Objective& Opening of cut & Optical\\
        Dicing width & [\SI{}{\micro\meter}]  & Objective& Opening of cut \& recast & Optical\\
        Recast & [\SI{}{\micro\meter}] & Objective& Height of recast above reference structure & Optical  \\ \midrule 
        Chipouts & [\%] & Constraint& Damage of top layer & Optical  \\
        Frontside cracks & [\%] & Constraint & Damage of top \& bulk material & Optical \\
        Frontside corner cracks & [\%] & Constraint & Damage in crossing & Optical \\
        Backside separation & [\%] & Constraint  & Separation yield & Optical \\
        Backside cracks & [\%] & Constraint & Cracks on backside of die & Optical \\
        Backside corner cracks &  [\%] & Constraint & Cracks on backside crossing & Optical \\
        \bottomrule
    \end{tabular}
    \end{small}
\end{table*}

\begin{table*}[t]
    \centering
    \begin{small}
    \caption{Expert-derived utility function weights used in both experiments and for the validation set analysis. To compute the die strength, we initially subtract a base value of $b=300$ prior to weighting it.}
    \label{tab:utility-weights}
    \begin{tabular}{lllllll}
    \toprule 
         & $w_1$: Dicing Width & $w_2$: Mod. Width & $w_3$: Burr height & $w_4$: Throughput & $w_5$: Front Str. & $w_6$: Back Str. \\ \midrule 
      Bare silicon wafer  & 0.075 & 0.075 & 1.0 & 0.01 & 0.5 & 0.5 \\ \midrule 
      Product wafer (\& BOLD-A)  & 0.05 & 0.05 & 0.1 & 0.3 & 0.25 & 0.25 \\   
      BOLD-B   &  0.005 & 0.005 & 0.01 & 0.098 & 0.45 & 0.45\\  
      BOLD-C   &  0.005 & 0.005 & 0.01 & 0.78 & 0.1 & 0.1\\    \bottomrule
    \end{tabular}
    \end{small}
\end{table*}

\section{Additional results}\label{app2}
\begin{figure}
    \centering
    \includegraphics[width=\linewidth]{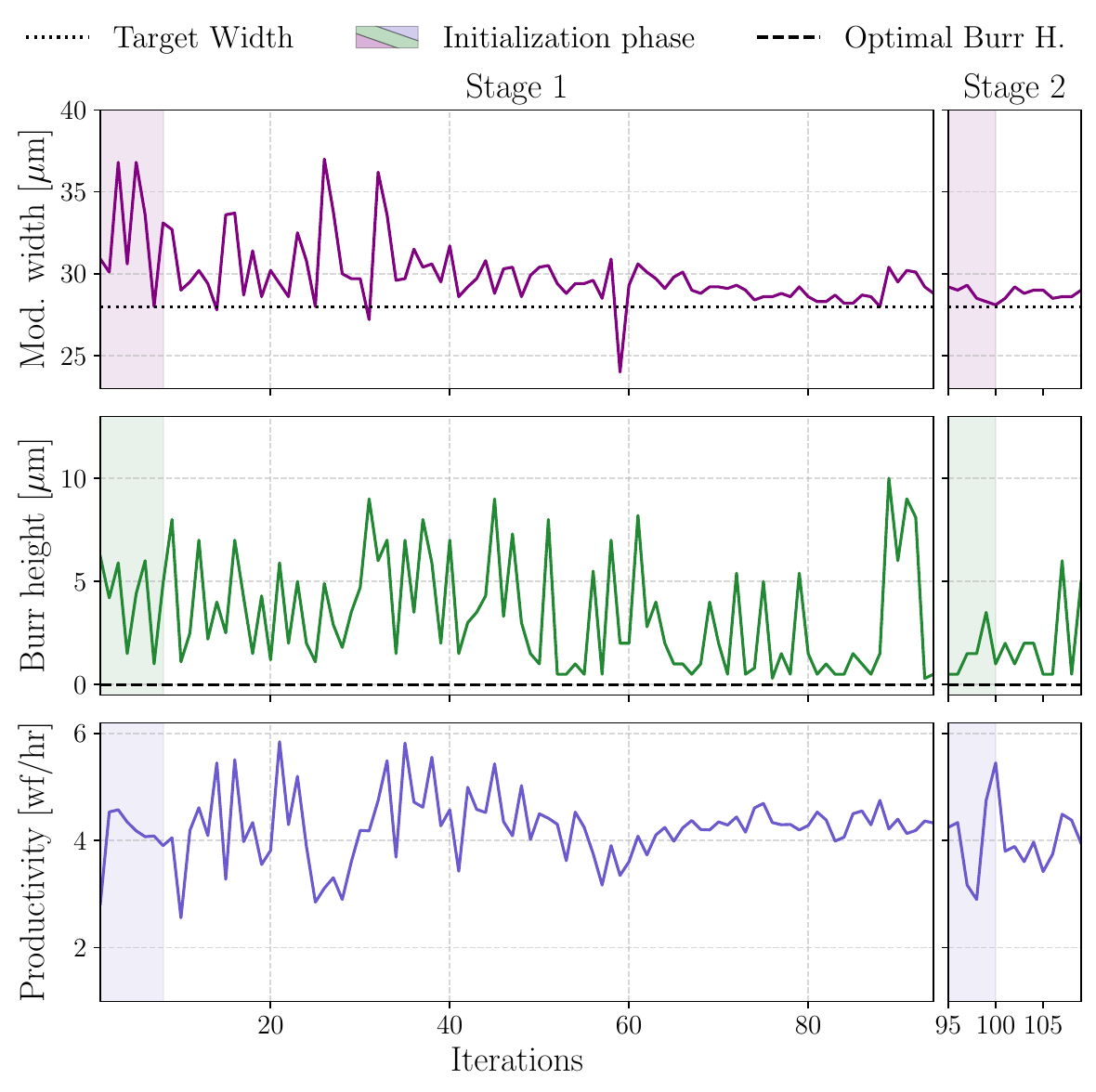}

    \caption{\textbf{Bare silicon wafer experiment convergence} - Per-iteration measurements for the three main objectives, visualizing the algorithm's exploration behavior. \textbf{(Top)} Modification width exploration, converging towards the 28 \si{\micro\meter} target. \textbf{(Middle)} Burr height exploration, showing a noisy trend of learning to avoid high-burr (bad) configurations over time. \textbf{(Bottom)} Productivity exploration, demonstrating the active search across the full range of process speeds.}
    \label{fig:experiment1-appendix}
\end{figure}

\begin{figure}
    \centering
    \includegraphics[width=\linewidth]{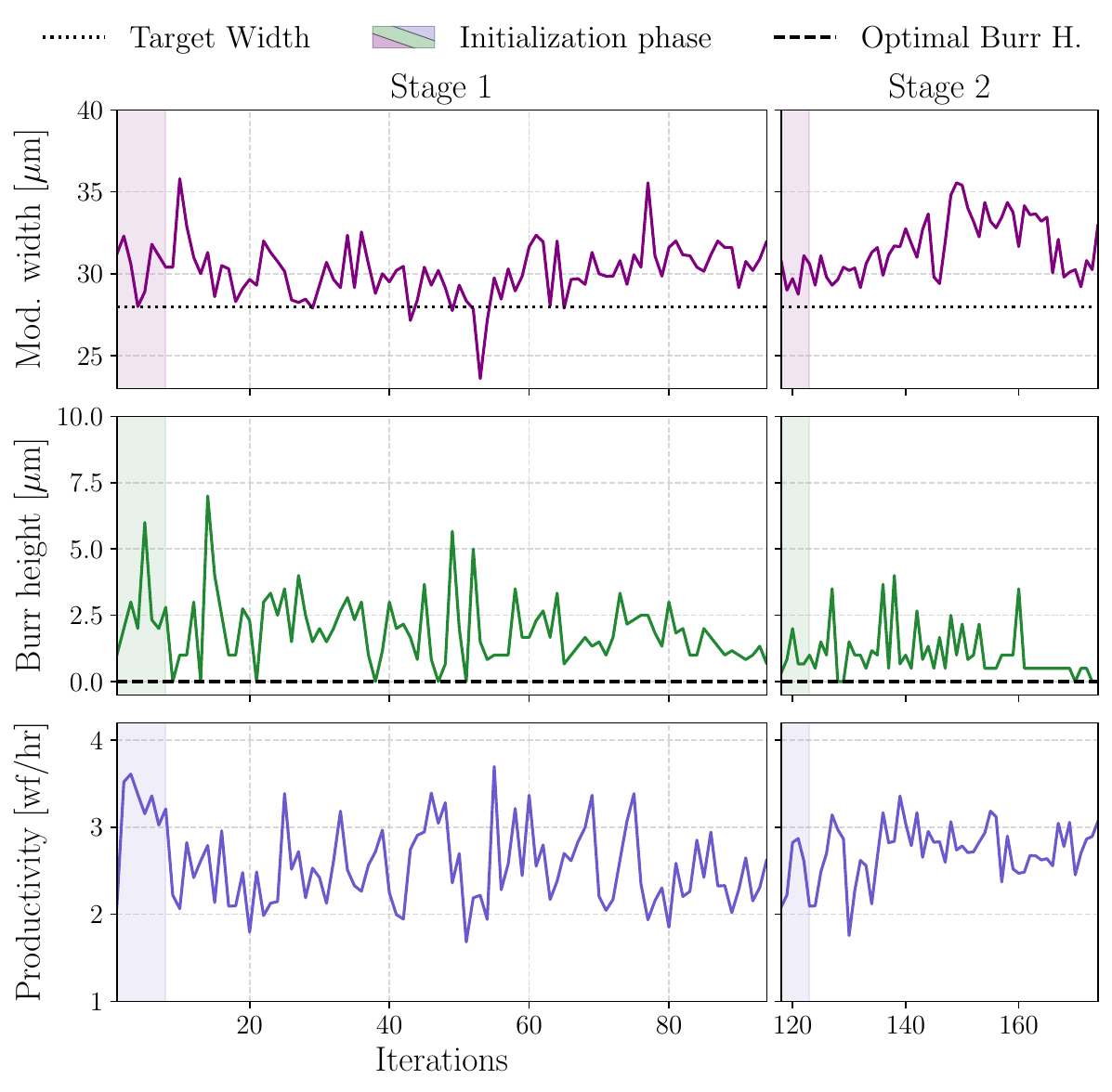}

    \caption{\textbf{Production wafer experiment convergence} - Per-iteration measurements for the three main objectives, visualizing the algorithm's exploration behavior. \textbf{(Top)} Modification width exploration, converging towards the 28 \si{\micro\meter} target. \textbf{(Middle)} Burr height exploration, showing a trend of learning to low-burr (good) configurations over time. \textbf{(Bottom)} Productivity exploration, demonstrating the active search across the full range of process speeds.}
    \label{fig:experiment2-appendix}
\end{figure}

To further analyze the optimization behavior, the convergence of individual objectives for both the bare silicon and production wafer experiments is visualized in Figs.~\ref{fig:experiment1-appendix} and~\ref{fig:experiment2-appendix}. In the bare silicon experiment, the model demonstrates strong convergence towards the ideal modification width, highlighting its ability to discover geometrically optimal recipes. In contrast, the production wafer experiment shows less aggressive convergence for this specific metric. This behavior is consistent with the expert-defined preferences (Table~\ref{tab:utility-weights}), where modification width was assigned a significantly lower utility weight. Instead, the optimization prioritizes burr height, resulting in a steadier convergence for this metric compared to the bare silicon case. Finally, the productivity traces reveal distinct exploration strategies. The bare silicon experiment exhibits high variance in the early stages as the model aggressively explores high-throughput regions; however, as these configurations often prove infeasible or detrimental to other objectives, the algorithm eventually stabilizes towards a robust, production-ready speed. Conversely, productivity in the production wafer experiment fluctuates throughout the entire run, likely a consequence of the stringent constraints making feasible high-speed configurations significantly harder to locate.

\newpage 

\bibliographystyle{elsarticle-harv} 

\bibliography{references.bib}
\end{document}